%% file: root.tex
\documentclass[letterpaper, 10 pt, conference]{ieeeconf}  % Comment this line out if you need a4paper

\IEEEoverridecommandlockouts                              % This command is only needed if 
\usepackage{hyperref}
\usepackage{graphicx} 
\usepackage{multirow}
\usepackage{caption}
\usepackage{algorithm}      
\usepackage{algpseudocode}
\usepackage{amsmath,amssymb} 
\usepackage[table,xcdraw]{xcolor}
\usepackage{stfloats}
\usepackage{booktabs}
\usepackage{makecell}

\newcommand{\highlightg}[2][B2F6B1]{%
  \begingroup
  \setlength{\fboxsep}{0pt}%
  \colorbox[HTML]{#1}{\strut #2}%
  \endgroup
}
\newcommand{\highlightr}[2][FFCCC9]{%
  \begingroup
  \setlength{\fboxsep}{0pt}%
  \colorbox[HTML]{#1}{\strut #2}%
  \endgroup
}

\title{\LARGE \bf
Toward Low-Latency Vision-Language Models with \\ Doubly-Correct Predictions in Egocentric Visual Understanding
}

\author{Qitong Wang$^{1, 2, *}$, Fan Du$^{1}$, Pranav Maneriker$^{1}$, Jihui Jin$^{1}$ and Christopher Rasmussen$^{2}$% <-this % stops a space
\thanks{*Work done during an internship at Dolby Laboratories, Inc.}% <-this % stops a space
\thanks{Corresponding author: wqtwjt@udel.edu}% <-this % stops a space
\thanks{$^{1}$The authors are with Dolby Laboratories, Inc.,
        1275 Market St, San Francisco, California, 94103, USA.
        {\tt\small Qitong.Wang@dolby.com; fan.du@dolby.com; Pranav.Maneriker@dolby.com; Jihui.Jin@dolby.com}}%
\thanks{$^{2}$The authors are with the Department of Computer and Information Sciences at the University of Delaware, Newark, DE 19711, USA.
        {\tt\small wqtwjt@udel.edu; cer@cis.udel.edu}}%
}
\begin{document}

\maketitle
\thispagestyle{empty}
\pagestyle{empty}

%%%%%%%%%%%%%%%%%%%%%%%%%%%%%%%%%%%%%%%%%%%%%%%%%%%%%%%%%%%%%%%%%%%%%%%%%%%%%%%%
\input{tex/0_abstract_0204}
\input{tex/1_introduction_0204}
\input{tex/2_related_works}
\input{tex/3_problem_formulation_3}
\input{tex/4_methodology}
\input{tex/5_experiment_0204}
\input{tex/6_conclusion}
\bibliographystyle{IEEEtran}
\bibliography{IEEEabrv, IEEEtranBST/IEEEfull}

\input{tex/x_appendix}

\end{document}

%% file: tex/0_abstract_0204.tex
\begin{abstract}

% Qitong on Feb. 4th, 2026
The rapid rise of Vision–Language Models (VLMs) in egocentric visual understanding has made low-latency inference in human-robot collaborative (HRC) tasks increasingly critical. 
Weight pruning techniques developed for VLMs to shrink model size and computation can be readily applied to satisfy the efficiency demands of on-board processing and real-time interactive robotics. 
Moreover, safe human-robot interaction demands pruning strategies that preserve doubly-correct predictions; outputs must be both accurate and evidentially grounded to mitigate risks and ensure user trust.
In this paper, we present a new study of VLM pruning through the lens of doubly-correct prediction. 
Our experiments surprisingly show that existing pruning methods often preserve the right evidence localization but undermine correct prediction. 
To address this, we propose a rationale-informed pruning strategy that better aligns evidence with decisions. 
Benchmark results on egocentric video datasets demonstrate that our method not only achieves the highest prediction accuracy but also outperforms existing approaches in attaining doubly-correct predictions.
We aim to stimulate research on efficient and reliable VLMs, ensuring accuracy-driven advances align with the transparency, auditability, and safety required for responsible human-robot interaction and embodied intelligence.

\end{abstract}

%% file: tex/1_introduction_0204.tex
\section{Introduction}

Vision–Language Models (VLMs)~\cite{clip, actionclip, internvideo} increasingly power robotic perception and decision-making systems, driving tasks ranging from visual navigation and manipulation to interactive agents in dynamic environments. 
Within this landscape, egocentric visual understanding stands out as a critical component for robotic autonomy~\cite{egoexo4d, ek55, ego4d}, as it aligns human and robot viewpoints to unlock skill transfer from diverse, real-world human activities~\cite{egovla}.
However, the remarkable performance of VLMs typically comes at the cost of massive parameter counts and heavy computational burdens. 
These overheads pose a severe challenge for robotic applications that mandate real-time responsiveness. 
For instance, in quadruped robot navigation, deploying high-capacity VLMs for semantic understanding on resource-constrained onboard hardware can easily result in prohibitive inference latencies—rendering the system far too slow for closed-loop control to prevent collisions or missed cues.
To mitigate this, model weight pruning~\cite{upop, multiflow, isomorphic, kwon2022fast, ecoflap} has emerged as a promising direction. 
By identifying and removing redundant computations, pruning effectively adapts massive VLMs to the strict low-latency constraints of egocentric robotic systems.

\begin{figure}[t]
  \centering
  \includegraphics[width=0.49\textwidth]{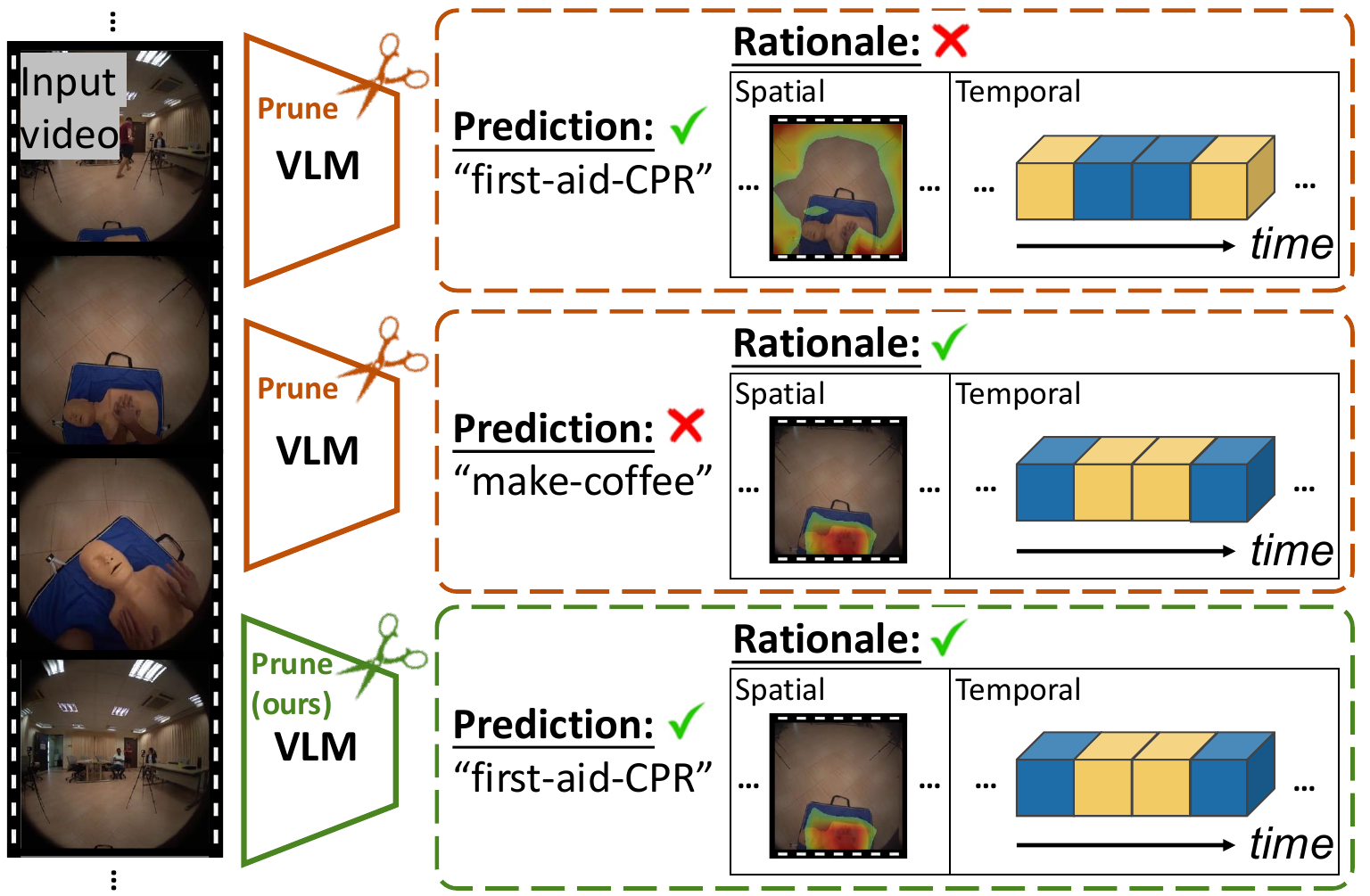}
  \caption{
  \textcolor[HTML]{4a8522}{Our method} departs from  \textcolor[HTML]{be5108}{existing weight pruning strategies} by ensuring not only correct prediction but also valid spatio-temporal rationales.
  Each cube in the temporal rationale represents a single time frame: \textcolor[HTML]{f2cb65}{yellow cubes} indicate frames containing the action of interest, while \textcolor[HTML]{1f6da8}{blue cubes} denote frames without the action of interest. 
  }
  % \vspace{-0.75cm}
\end{figure}

However, most existing weight pruning methods optimize solely for prediction accuracy, inadvertently neglecting the critical property of \textit{doubly-correct predictions (DCP): predictions that are not only factually correct but also grounded in valid visual evidence}~\cite{mao2023doubly, li_rationale, wang_rationale, ma_rationale}. 
While acceptable in pure recognition tasks, overlooking prediction groundedness poses severe safety risks in physical robotic deployment. 
Consider a human-robot collaboration scenario where a VLM guides a robot to ``hand over the wrench." 
A model optimizing only for accuracy might correctly predict the intent (``pass wrench") but base this decision on spurious correlations, such as a specific background texture, rather than the wrench itself. 
% Crucially, robotic control often relies on the model's visual attention to localize the object for grasping. 
% If the model's reasoning is misaligned (i.e., focusing on the background), 
This misalignment may cause the robot to grasp empty space or a bystander object, despite the correct high-level prediction. 
Such ``Right for the Wrong Reason" failures can lead to handing over the wrong tool, moving unpredictably, task failure, human injury, or a breakdown of trust in the system, undermining the transparency and reliability for robotics. 
Therefore, our core research question is:
\textit{How does weight pruning affect the safety of VLM inference? 
Can we design a method that reduces computation costs while preserving the model's ability to produce doubly-correct predictions?}

% To the best of our knowledge, this is the first work to investigate this problem. 
We begin by defining doubly-correct prediction evaluation in the context of egocentric visual understanding. 
Based on our evaluation, we apply existing weight pruning methods to standard benchmarks and uncover new findings.
We observe that current pruning methods often lead to samples with wrong predictions but correct rationales: when focusing on the valid evidence of target objects, pruned VLMs are less likely to produce correct predictions. 
% In other words, although these models can accurately localize the relevant evidence, that evidence does not necessarily enable them to make the right prediction.
In other words, such valid evidence often fails to lead to the correct prediction.

Building on the above finding, we propose a new rationale-informed pruning method. 
The goal is to ensure that when a model identifies correct rationales, it can more effectively leverage this evidence to support accurate predictions. 
Our method leverages the unequal importance of parameters across layers by assigning non-uniform pruning ratios based on weight magnitudes. 
Unlike prior approaches that rely solely on activations, we exploit the model’s own rationales to guide pruning, thereby preserving evidence that is truly discriminative and semantically relevant to the prediction of VLMs.
This rationale-informed strategy ensures that pruning not only retains correct rationales but also enables accurate predictions, ultimately promoting doubly-correct prediction.
Experiments on standard benchmarks demonstrate that our method surpasses existing pruning approaches in both prediction accuracy and doubly-correct prediction. 
By focusing on valid rationales, our method enables more effective alignment between rationale and prediction, thereby improving overall accuracy and the proportion of doubly-correct samples.

The contributions of this paper are as follows:

% $\bullet$ We propose a new evaluation protocol for egocentric visual understanding and understanding, termed spatio-temporal doubly-correct prediction evaluation.
$\bullet$ We propose a spatio-temporal doubly-correct evaluation protocol for egocentric visual understanding.

$\bullet$ Through experiments with existing pruning methods, we uncover new phenomena showing that current pruning strategies lead to unreliable VLM deployment.

$\bullet$ We introduce a new pruning approach that effectively preserves VLMs' doubly-correct prediction ability while reducing parameters. 
Experiments on standard benchmark datasets demonstrate the effectiveness of our method.

%% file: tex/2_related_works.tex
\section{Related Works}

% \subsection{Vision-Language Models for Video Learning}

% Recently, Vision-Language Models (VLMs) have attracted significant attention. 
% In video-related tasks, models such as ActionCLIP~\cite{actionclip} 
% have demonstrated strong predictive performance in both zero-shot and fine-tuned settings, across a wide range of tasks, including action recognition, temporal action localization, video retrieval, and video question answering. 
% This is due to their large-scale pretraining on vision–language pairs, which equips them with strong cross-modal generalization. This enables robust performance on diverse video tasks with limited task-specific supervision.
% However, these advantages come at the cost of massive parameter counts, which substantially increase computational overhead and latency in real-world deployment.

\subsection{Vision-Language Models for Egocentric Understanding}

Egocentric visual understanding is a core problem in robotics and embodied AI, where first-person observations capture agent-centric interactions with the environment. 
Built on the CLIP-style~\cite{clip} vision-language pretraining paradigm, existing works including~\cite{actionclip, egovlpv2} provide a strong and practical foundation for dynamic vision representation learning, with solid transferability in both zero-shot and fine-tuned settings. 
% In this work, we therefore take CLIP/ActionCLIP as the primary VLM backbone for studying egocentric visual understanding. 
% Recent egocentric-focused VLMs (e.g., EgoVLP, LaViLa, and EgoVLPv2) further demonstrate that large-scale video-language pretraining can improve first-person benchmarks, which is complementary to---rather than replacing---the CLIP/ActionCLIP line. 
This is due to their large-scale pretraining on vision--language pairs, which equips them with strong cross-modal generalization.
This enables robust performance on diverse tasks with limited task-specific supervision.
However, these advantages come at the cost of massive parameter counts, which substantially increase computational overhead and latency in real-world deployment.

\subsection{Transformer Weight Pruning}
To address the above-mentioned challenge, researchers have extensively explored model pruning as a promising solution.
One common approach~\cite{upop, multiflow, isomorphic} is to prune a dense model and then fine-tune it on the target task. 
A recent example is UPop~\cite{upop}, which jointly searches multimodal subnets in a continuous space with automatic pruning ratio assignment and progressively retrains them to ensure convergence and achieve higher compression. 
MULTIFLOW~\cite{multiflow} prunes VLMs once while preserving transferability to unseen downstream tasks by leveraging parameter magnitude and information flow. 
% However, the abovementioned methods need retraining after pruning, which often incurs substantial training costs and increases deployment complexity, limiting practical usability. 
However, these methods require retraining after pruning, incurring substantial computational cost and deployment complexity, which limits their practical usability.
To address this, another line of work explores pruning without any retraining. 
For instance,~\cite{kwon2022fast} employs a lightweight Fisher information–based mask search to identify prunable heads and filters, followed by mask rearrangement and tuning to refine layer-wise activations. 
Nevertheless, these methods overlook the rationale behind model predictions. 
Thus, how to prune models while preserving doubly-correct predictions remains an open and important problem.

\begin{figure*}[t]
\vspace*{0.05in}
  \centering
  \includegraphics[width=1.0\textwidth]{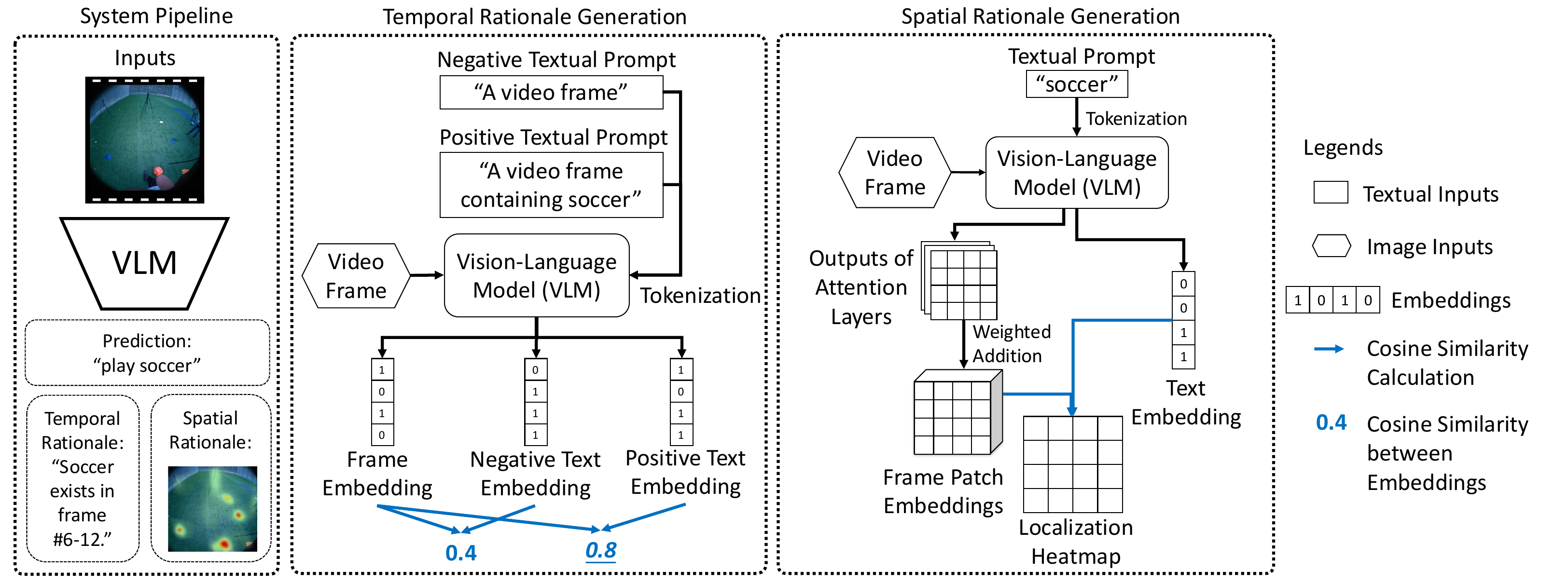}
  \caption{
  Our system takes an egocentric video clip as input and feeds it into a Vision–Language Model (VLM). 
  The VLM produces not only a prediction, such as identifying the action performed by the person in the video, but also a spatio-temporal rationale for this prediction. 
  Specifically, the temporal rationale determines in which frame(s) the rationale object occurs, while the spatial rationale localizes the corresponding object within those frames.
  We cast temporal rationale generation as a binary classification per frame, where the VLM receives the frame and two text prompts, following the CLIP-style prompting strategy~\cite{clip}.
  For spatial rationale generation, we derive an explanation heatmap by extracting attention head outputs and computing cosine similarity between patch tokens and text tokens, following the method in~\cite{li_rationale}.
  }
  \label{fig_rationale_eval}
\end{figure*}

\subsection{Doubly-Correct Predictions in VLMs}
For vision-language models (VLMs), the concept of Doubly-Correct (or Doubly-Right) was first introduced by~\cite{mao2023doubly}. 
It emphasizes that VLMs' predictions should be correct not only in label but also grounded in valid evidence—that is, supported by correct rationales.
Specifically, \cite{mao2023doubly} proposed a ``why-prompt'' approach that enables CLIP to retrieve both the right category and the rationales justifying its textual prediction, given a visual input.
Recently, several studies have explored visual rationales in VLM-based image understanding. 
\cite{li_rationale} introduced a rationale-informed optimization method to improve doubly-correct predictions. 
\cite{wang_rationale} proposed new evaluation metrics from the perspective of doubly-correct predictions and examined how fine-tuning influences the rationality of VLMs' predictions. 
In the medical domain, \cite{ma_rationale} curated a rationale dataset and designed a rationale-informed optimization strategy that disentangles and localizes fine-grained clinical concepts, thereby extending doubly-correct predictions to medical AI.
However, ensuring doubly-correct predictions in egocentric visual understanding remains under-explored. 
The unique challenges of egocentric visual understanding, such as spatio-temporal characteristics of vision data, and the demand for low-latency processing, distinguish it from image and text tasks, thereby limiting the applicability of existing studies.

%% file: tex/3_problem_formulation_3.tex
\section{Problem Formulation}

We investigate doubly-correct predictions in egocentric visual learning and understanding, a setting that has received limited attention in prior work.
Egocentric visual understanding introduces additional challenges, including spatio-temporal rationale requirements and latency-oriented deployment constraints.
Therefore, while existing image- and text-based studies provide valuable foundations, they cannot be directly applied without adapting the evaluation protocol to account for both \textit{where} and \textit{when} visual evidence appears.

\subsection{Overview of the Evaluation Pipeline}
\label{sec_3_0}

The goal of our evaluation is to determine whether a VLM is correct in both prediction and rationale.
Given an egocentric video, we first evaluate whether the VLM predicts the correct action label.
We then evaluate whether its rationale correctly identifies the manipulated object in both the temporal dimension, i.e., \textit{when} the object appears, and the spatial dimension, i.e., \textit{where} the object appears.
Finally, we combine prediction correctness and rationale correctness to compute doubly-correct prediction (DCP) metrics.
After defining this protocol in Sections~\ref{sec_3_1}--\ref{sec_3_3}, we apply it to diagnose existing pruning methods in Section~\ref{sec_3_4}.

\subsection{Task Setup and Egocentric Rationale Definition}
\label{sec_3_1}

\textbf{Prediction Evaluation.}
There has been growing interest in applying VLMs to egocentric vision downstream tasks, with action recognition being a typical example.
We focus on video classification tasks~\cite{egoexo4d, ekv} and assess VLM's performance using the top-1 accuracy metric.

\textbf{Rationale Definition.}
We define the rationale in egocentric video action recognition as \textit{the manipulated object that is distinctively associated with the human action}.
We adopt this definition based on the ``Sufficient Input Subsets'' (SIS) theory~\cite{sis}, which posits that the smallest input subset preserving the model's prediction constitutes a sufficient explanation.
In egocentric vision, actions are typically composed of a verb and a noun~\cite{ek55, egaze}.
To identify the minimal yet sufficient rationale from these components, we analyze them as follows:
(1) \textbf{Ambiguity of Body Parts}: While human hands are ubiquitous in egocentric videos, they lack action specificity, as hand shapes for distinct actions like ``cutting'' or ``washing'' can appear visually similar.
(2) \textbf{Ambiguity of Verbs}~\cite{verb_ambi}: Verb labels often exhibit semantic ambiguity; for instance, the same action of processing a carrot could be annotated as ``cut'' or ``peel''.
(3) \textbf{Primacy of Objects}: Cognitive science findings~\cite{n_1st} suggest that humans learn to prioritize identifying objects over verbs during action recognition.
Consequently, we contend that the \underline{human-manipulated objects} serve as the most robust and sufficient rationale for egocentric action recognition.

\textbf{Definition 1} (Rationale in Egocentric Video).
\textit{Rationales are defined as human-manipulated objects that provide explicit evidence for the model's predictions.}

\subsection{Spatial and Temporal Rationale Correctness}
\label{sec_3_2}

In contrast to still images, egocentric videos intrinsically encode spatio-temporal dependencies~\cite{lzy_st, st2}, linking \textit{where} actions appear in frames (spatial) with \textit{when} they occur over time (temporal).
Accordingly, we define the rationale for egocentric video learning along two dimensions: spatial (i.e., where) and temporal (i.e., when).

\textbf{Definition 2.1} (Spatial Rationale).
\textit{Spatial rationale is defined as the explanation of human-manipulated objects, namely their localized positions within frames.}

\textbf{Definition 2.2} (Temporal Rationale).
\textit{Temporal rationale is defined as the per-frame presence of manipulated objects.}

The overall evaluation framework is shown in Figure~\ref{fig_rationale_eval}.
We evaluate with two steps.
(1) \textbf{Temporal Rationale}:
The VLMs must determine in which frame(s) of the input video clip the rationale object appears.
We formulate this as a binary classification problem for each frame, where the VLMs are given the frame together with two text prompts (e.g., [``a video frame'', ``a video frame containing soccer'']), following the CLIP-based prompting strategy~\cite{clip}.
(2) \textbf{Spatial Rationale}:
For frames containing the target object, we evaluate spatial correctness at the object level by verifying whether the VLM's explanation heatmap strongly activates on the corresponding object regions.
Therefore, following~\cite{wang_rationale}, we adopt Relevant Mass Accuracy (RMA)~\cite{es_eval1, es_eval2} rather than the commonly used Intersection over Union (IoU).
RMA is calculated by determining the ratio of the total heatmap pixel values within the target object regions to the sum of all pixel values across the entire heatmap.
It requires both the generated explanation heatmap $(H)$ from VLMs and the ground truth explanation mask $(M)$, whose pixels on the target objects are marked as 1, otherwise marked as 0.
RMA score is defined as:
\begin{equation}
\text{RMA}(H, M) = \frac{\sum H \odot M}{\sum H},
\label{es_eval}
\end{equation}
where $\odot$ represents the Hadamard product.

The standard RMA score penalizes background activations, thereby functioning as a precision-oriented metric (denoted hereafter as $RMA_p$).
To simultaneously account for recall, specifically, the coverage of the target object region, we introduce a variant, $RMA_r$.
While $RMA_p$ normalizes by the total heatmap mass $\sum H$, $RMA_r$ normalizes by the ground-truth mask mass $\sum M$.
To evaluate the trade-off between precision and recall, we adopt the weighted F-score:
\begin{equation}RMA_f = \frac{(1 + \beta^2) \cdot RMA_p \cdot RMA_r}{\beta^2 \cdot RMA_p + RMA_r}.\label{eq:rma_f}\end{equation}
While $\beta=1$ (the F1 score) is the standard choice for balanced evaluation, we argue that recall should be prioritized in egocentric spatial rationale evaluation.
This decision is grounded in the asymmetry of error costs:
(1) \textbf{Impact on Model Performance}:
Qualitative studies~\cite{rma_f2} indicate that missing key objects (false negatives) is a primary cause of prediction failure.
In contrast, rationales that cover the object but include some background noise (false positives) often preserve correct predictions.
(2) \textbf{Impact on Shared Autonomy}:
In teleoperated grasping, if the visual rationale fails to highlight the target object's affordance (false negative), the human operator may miss the grasp entirely.
Conversely, a visualization that covers the target but spills over onto the background (false positive) remains informative, as the operator can intuitively filter out the noise and focus on the graspable area.
Given that false negatives are significantly more detrimental than false positives, we set $\beta = 10$.
We note that $\beta$ is used only for evaluation, not for pruning, and the same criterion is applied uniformly to all compared methods.
Finally, we binarize the evaluation by thresholding $RMA_f$ at 0.5, where scores $\ge 0.5$ are considered correct.

\subsection{Doubly-Correct Prediction Metrics}
\label{sec_3_3}

Our doubly-correct prediction (DCP) evaluation protocol accounts for the relations between prediction and rationale correctness.
Therefore, we adopt the evaluation procedure proposed in~\cite{wang_rationale}.
Specifically, VLM's outputs are categorized into four scenarios based on prediction accuracy (right/wrong) and rationale validity (right/wrong), denoted as RR (right prediction \& right rationale), RW (right prediction \& wrong rationale), WR (wrong prediction \& right rationale), and WW (wrong prediction \& wrong rationale).
From these, two complementary metrics are defined: Prediction Trustworthiness (PT), measuring the proportion of valid rationales among correct predictions, defined as $\frac{RR}{RR + RW}$; and Inference Reliability (IR), measuring the proportion of correct predictions among valid rationales, defined as $\frac{RR}{RR + WR}$.
Last but not least, we also measure the proportion of $RR$ among all samples (defined as $\frac{RR}{RR + RW + WR + WW}$), where RR represents the ideal cases.
We evaluate spatial and temporal DCP separately, so that both spatial and temporal PT/IR/RR are reported.
These provide a comprehensive assessment of VLM's prediction safety across multiple perspectives.
Higher scores indicate better performance.

\subsection{Diagnostic Study of Existing Weight Pruning Methods}
\label{sec_3_4}

Having defined the spatio-temporal DCP protocol, we next use it to diagnose how existing post-training pruning methods affect VLM reliability.
We focus on post-training weight pruning because retraining-based pruning can introduce additional deployment complexity and computational cost, limiting its feasibility for resource-constrained embodied systems~\cite{kwon2022fast}.
We investigate four representative methods: one-shot magnitude pruning (OMP)~\cite{omp}, global magnitude pruning (GMP)~\cite{gmp}, Kwon et al.~\cite{kwon2022fast}, and ECoFLaP~\cite{ecoflap}.

% Given our focus on egocentric learning, w
We select two representative benchmarks: EPIC-KITCHENS VISOR~\cite{ekv} and EgoExo4D~\cite{egoexo4d}.
Both datasets provide annotations for model predictions and ground-truth rationales, such as pixel-level masks of manipulated objects, which are required by our spatio-temporal DCP evaluation.
We use ActionCLIP~\cite{actionclip} (ViT-B/32) as our canonical model because it is an encoder-only VLM for video understanding with well-defined classification outputs and accessible visual representations for rationale evaluation.
Although video MLLMs~\cite{videollava, videollama3}, and VLAs~\cite{navila, streamvln}, are advancing rapidly, their open-ended outputs and action-generation interfaces introduce distinct rationale evaluation challenges.
We leave their evaluation as a future work.
For spatial rationale generation, we adopt the method proposed by~\cite{li_rationale}, which computes the cosine similarity between output patch tokens and text tokens to generate a heatmap as the spatial rationale, following prior rationale evaluation studies.
For pruning calibration~\cite{ecoflap}, we uniformly sample 20\% of the data from the evaluation set according to the index order, while the remaining samples are used for testing.

Before pruning VLMs, we fine-tune all the VLM's parameters until convergence, adopting the default fine-tuning setups of the VLMs.
To evaluate computational cost, we use Floating-Point Operations Per Second (FLOPs), a hardware-independent proxy widely adopted in prior pruning work.
However, actual wall-clock latency also depends on hardware, memory access, kernel implementation, and deployment framework.
Each result is run once with a fixed seed and a deterministic setup, and repeated runs under the same settings yield identical results.
In contrast to image classification pruning settings, where existing work~\cite{ecoflap} sets the pruning ratio up to 60\%, our pruning ratio for video tasks is set lower, between 20\% and 30\%, because video classification exhibits a noticeable performance drop even at relatively low pruning ratios.

\input{tabs/flops}

The VLM we adopt consists of three major components: the vision encoder, the text encoder, and the multimodal fusion module.
From Table~\ref{component_gflop}, we observe that the vision encoder accounts for the largest portion of the computational cost.
In particular, the computation of the vision encoder is around 2 times that of the text encoder and around 37 times that of the multimodal fusion module.
Therefore, to strike a trade-off between prediction performance and efficiency, we prune only the vision encoder in all subsequent experiments, while keeping the other components unchanged.

Surprisingly, as observed in Tables~\ref{tab_res_actionclip32_ekv} and \ref{tab_res_actionclip32_ee4d}, \textit{while existing methods (OMP, GMP, Kwon et al., and ECoFLaP) maintain a high proportion of valid rationales among correct predictions, they suffer from two key drawbacks: they reduce the absolute number of doubly-correct instances, and they increase the frequency of incorrect predictions accompanied by valid rationale}.
As expected, the prediction accuracy decreases.
Notably, our analysis reveals a nuanced divergence in how pruning affects model safety.
On the surface, methods like ECoFLaP demonstrate surprising resilience regarding prediction trustworthiness; at a 20\% pruning ratio (Table~\ref{tab_res_actionclip32_ekv}), ECoFLaP actually yields a 4.81\% improvement in spatial PT scores compared to the unpruned baseline, suggesting that pruning does not inherently decouple correct predictions from valid regions.
However, despite this apparent stability, a deeper examination exposes a significant degradation from the perspective of doubly-correct predictions.
We observe substantial declines in both IR and RR scores—specifically an 11.09\% drop in spatial IR and a 2.45\% decrease in spatial RR for the same model.
This indicates that while the pruned model retains the capacity to identify valid rationales, it fails to effectively leverage these cues to drive correct predictions.
Our findings challenge the prevailing assumption that pruning only affects efficiency and accuracy, and instead uncover a previously unexplored dimension of risk of pruning.

%% file: tabs/flops.tex
\begin{table}[h]
\small
% \footnotesize
\centering
\caption{The GFLOPs (Giga FLOPs) analysis of different model components of ActionCLIP clearly shows that the vision encoder dominates the computational cost.}
% \begin{tabular}{cccc}
% \hline
% \begin{tabular}[c]{@{}c@{}}Vision Encoder \\ ViT-B/16\end{tabular} & \begin{tabular}[c]{@{}c@{}}Vision Encoder \\ ViT-B/32\end{tabular} & \begin{tabular}[c]{@{}c@{}}Text \\ Encoder\end{tabular} & \begin{tabular}[c]{@{}c@{}}Multimodal \\ Fusion Model\end{tabular} \\ \hline
% 24.159                                                             & 7.512                                                              & 3.876                                                   & 0.201                                                              \\ \hline
% \end{tabular}

\begin{tabular}{@{\hspace{10pt}}c@{\hspace{28pt}}c@{\hspace{28pt}}c@{\hspace{10pt}}}
\hline
\begin{tabular}[c]{@{}c@{}}Vision Encoder \\ ViT-B/32\end{tabular} &
\begin{tabular}[c]{@{}c@{}}Text \\ Encoder\end{tabular} &
\begin{tabular}[c]{@{}c@{}}Multimodal \\ Fusion Model\end{tabular} \\
\hline
7.512 & 3.876 & 0.201 \\
\hline
\end{tabular}
\label{component_gflop}
\end{table}

%% file: tex/4_methodology.tex
\section{Methodology}
% Building on the findings in Section~\ref{sec_3_2}, we propose a new post-training pruning method with two key objectives:
% (1) Preserve the strengths of existing pruning methods by avoiding degradation of the original PT score of VLMs.
% (2) Leverage VLMs’ rationale localization to guide more accurate predictions and improve doubly-correct performance over existing methods.

Building on the findings in Section~\ref{sec_3_2}, we propose a novel post-training weight pruning framework. 
While maintaining the same pruning ratio and a similar level of computational overhead (FLOPs) as existing baselines, we target two objectives: 
(1) \textbf{Preserve Trustworthiness}: 
Maintain the Prediction Trustworthiness (PT) scores characteristic of current state-of-the-art methods. 
(2) \textbf{Bridge the Rationale-Prediction Gap}: 
Effectively leverage the model's rationale localization to drive accurate decision-making, thereby significantly improving \textbf{Doubly-Correct} performance.

Motivated by~\cite{textspan1, textspan2}, which show that layers in VLMs' vision encoder contribute unequally to prediction, we posit that parameters at different depths have different importance. 
% Therefore, under a fixed overall pruning budget, the per-layer pruning ratios should be non-uniform. 
% As a first step, we estimate each layer’s pruning ratio based on its magnitude to guide how much to prune from each layer.
Therefore, under a fixed overall pruning budget, the per-layer pruning ratios should be inherently non-uniform. 
As a first step, we roughly estimate each layer’s pruning ratio based on the magnitude of its weights to better guide how much to prune from each layer.

\begin{algorithm}[h]
\caption{Rationale-Informed Pruning}
\begin{algorithmic}[1]
\State \textbf{Input:} Weights $\{W_l\}$, input data activations $X$, global pruning ratio $\rho$, damping $\lambda$, rationale masks $\{O'\}$, all-ones matrix $I'$.
\State \textbf{Output:} Pruned weights $\{\widehat{W}_l\}$
% \Statex
\For{$l=1$ to $L$}
    \State Compute weight magnitude score $R_l \gets \|W_l\|_1$
    % \State Compute weight magnitude score $R_l \gets \|W_l\|_1 \cdot \|\nabla_{W_l}\|_1$
    \State Allocate per-layer pruning budget $\rho_l$
    % \State Compute text similarity $c$ of the explanation object and the predicted class and modulate mask $O \gets c O' + (1-c)I'$
    \State Compute the action–rationale text similarity $c$
    \State Modulate the mask as $O \gets c O' + (1-c)I'$
    \State Build the Hessian matrix $H_{jj}(l) \gets (X_j(l) \odot O)(X_j(l)^T \odot O) + \lambda I$
    \State Score weights $\varepsilon_j \gets \dfrac{||W_j(l)||^2}{\text{diag}(H_{jj}(l))^{-1}}$
    \State Keep top-$k_l$ weights based on $\varepsilon_j$ to form $\widehat{W}_l$
\EndFor

\end{algorithmic}
\label{alg_ours}
\end{algorithm}

\input{tabs/res_actionclip32_ekv}
Specifically, for the weights in a given layer \( l \), we determine its pruning ratio ($R_l$) based on the magnitude-weighted L1 norm: $\left\| W_l \right\|_1$.
After determining the pruning ratio for each layer, we assign an importance score to every weight in that layer. 
We then preserve the weights with higher scores and prune those with lower scores.
A growing line of work prunes neural networks by combining weight (denoted as $W$) magnitudes with data-dependent statistics of the input activations (denoted as $X$) to estimate parameter importance. 
% Among these, SparseGPT~\cite{sparsegpt} is a representative state-of-the-art layer-wise pruning method that leverages row-wise Hessian reconstruction for accurate weight importance estimation. It has been widely adopted in recent works as a strong baseline for large-scale model pruning.
% It prunes layer weights via a sparse regression step that, for each weight row, reconstructs the layer’s output using a row-wise Hessian (from $H = XX^T + \lambda I$) and removes parameters with the smallest impact, based on Optimal Brain Surgeon (OBS)-style~\cite{obs} importance ($\varepsilon \;=\; {||W||^{2}}/diag{\big(H^{-1}\big)}$).
% A broad family of post-training, layer-wise pruning methods scores parameters by combining weight magnitudes with data-dependent input activations. 
% Representative state-of-the-art include Wanda~\cite{wanda}, which uses activation-aware magnitude ranking, and SparseGPT~\cite{sparsegpt}, which operates via weight updates, aiming to preserve the input–output relation at each layer.
Representative state-of-the-art include Wanda~\cite{wanda}, which uses activation-aware magnitude ranking, and SparseGPT~\cite{sparsegpt}, which operates via weight updates to preserve the input–output relation at each layer. 
Both methods also serve as key components within the existing method, such as ECoFLaP~\cite{ecoflap}, forming the basis upon which our pruning strategy is built.

In VLMs, pruning criteria that aggregate input activations across spatial and temporal locations inevitably capture a substantial amount of background information.
From the spatial perspective, the feedforward layers in Transformers 
% (following the attention layers) 
can be regarded as key–value memories that transform attended rationales into feature components useful for classification~\cite{geva2021transformer}. 
We posit that pruning that disproportionately emphasizes background activations risks removing these critical ``memory units/channels,'' thereby breaking the transmission from rationale to logits: attention maps may still highlight the correct spatial regions, but the fine-grained or temporal components aligned with the class fail to propagate, resulting in incorrect predictions (WR). 
This reasoning extends naturally to temporal rationales, which we define as frame-level binary decisions of object presence. 
Since these rationales are classification outcomes derived after passing through all model components, pruning that over-weights background frames might impair the propagation of discriminative evidence across frames. 
Consequently, rationales may remain valid at the frame level, yet the aggregated video-level prediction becomes incorrect. 
% (WR).

Motivated by this, we propose to leverage the model’s own rationales during pruning to mitigate the influence of background activations when computing importance scores. 
The key advantage is that our approach requires no additional supervision: once the model has already produced valid rationales, these rationales can be directly exploited to guide pruning, preserving evidence that is truly discriminative. In this way, pruning not only retains correct rationales but also enables the model to make correct predictions, thereby achieving the goal of doubly-correct prediction.

Building on OBS~\cite{obs} as our theoretical foundation, we express the pruning-induced loss increase as a computable quadratic form so that weight importance can be obtained. 
To estimate the loss rise caused by deleting weight, we use its second-order Taylor approximation:
\begin{equation}
% {
% \begin{split}
%     \Delta L &= L\bigl(W + \delta W\bigr) - L\bigl(W\bigr) \\
%              &\approx \tfrac12\,\delta W^\top H\,\delta W,
% \end{split}
% }
\Delta L \approx \tfrac12\,\delta W^\top H\,\delta W,
\label{second_order_formula1}
\end{equation}
where $H$ is the Hessian matrix, that defined as the second derivatives of the loss with respect to the parameters, characterizes the model's local curvature around the current parameters.
OBS obtains a closed-form solution for formula~\ref{second_order_formula1} by solving a constrained quadratic minimization problem:
\begin{equation}
% \begin{aligned}
\min_{\delta W}\; \tfrac{1}{2}\,\delta W^{\top} H\,\delta W \qquad
\text{s.t.}\; e_{j}^{\top}\delta W = -\,W_{j}\,.
% \end{aligned}
\label{second_order_formula2}
\end{equation}
Solving the Lagrangian of formula~\ref{second_order_formula2} yields the optimal-cost importance score ($\varepsilon_j$) for each weight ($W_j$):
\begin{equation}
% \small
    \varepsilon_j = \,\frac{||W_j||^{2}}{diag\big(H_{jj}\big)^{-1}} \ = \,\frac{||W_j||^{2}}{diag\big((X_j \odot O)(X_j^T \odot O) + \lambda I\big)^{-1}}, 
    % \[
    %  \;\propto\; \frac{w_j^2}{\big(H^{-1}\big)_{jj}}
    % \]
\end{equation}
% where $X_j$ represents the input data activation that participates in the inference of weight ($W_j$), and $O$ is rationale mask, and $\lambda$ is a damping term and $I$ is an Identity matrix to ensure numerical stability.
where $X_j$ denotes the input activation associated with weight $W_j$, 
$O$ is the rationale mask, 
$\lambda$ is a damping term, 
and $I$ is the identity matrix introduced to ensure numerical stability.
As argued earlier, the Hessian matrix $H$ should not be built from raw inputs ($X_j$) alone because a substantial portion corresponds to background.

\input{tabs/res_actionclip32_ee4d}

% We therefore weight based on the rationale mask $O$ and write the $H$ as:
% \begin{equation}
%     H = (XO)(O^TX^T) + \lambda I,
% \end{equation}

Using the rationale mask as $O$ is suboptimal, as the mask itself is noisy.
When the object description is more semantically aligned with the predicted class, the contribution inherited from the original mask $O'$ should be amplified; otherwise, it should be down-weighted.
We compute the cosine similarity (denoted as $c$) between their text embeddings of the explanation object and the predicted class, and use it as a semantic factor. 
Accordingly, we define the refined mask $O$ as a convex combination of the original mask $O'$ and an all-ones baseline matrix $I'$ with the identical shape:
\begin{equation}
    % O = c * O' + (1-c) * I',
    O \;=\; c\, O' \;+\; (1-c)\, I', 
    % \qquad
    % O',I'\in\mathbb{R}^{n\times m},
    \quad c\in[0,1].
\end{equation}
% where $I'$ is an all-ones matrix with the same size (dimensions/shape) as $O'$.
% We choose $c$ as 0.5 in our experiments.
Our pseudocode is provided in Algorithm~\ref{alg_ours}. 
% Notably, our method is also a post-training pruning approach.
We next conduct experiments to demonstrate our method’s effectiveness.

%% file: tabs/res_actionclip32_ekv.tex
\begin{table*}[t]
\vspace*{0.1in}
\centering
\caption{Experimental results with the ActionCLIP ViT-B/32 model on the EPIC-KITCHENS VISOR dataset.
In each setup, for prediction accuracy (Pred. Acc.) as well as spatial and temporal IR and RR, we display the first-place results in \textbf{bold}.
In the spatial and temporal PT columns, scores higher than the dense baseline are highlighted in \highlightg{green}, while the others are highlighted in \highlightr{red}.
``P-Ratio'' denotes the pruning ratio of the vision encoder of VLMs.
}
\label{tab_res_actionclip32_ekv}
{\small
% {
\setlength{\tabcolsep}{1pt}
\renewcommand{\arraystretch}{1.0}

\begin{tabular*}{\textwidth}{@{\extracolsep{\fill}} c c c c c c c c c c c c c c}
\hline
\multirow{2}{*}{Pruning} & \multirow{2}{*}{P-Ratio (\%)} & \multirow{2}{*}{GFLOPs} &  & \multicolumn{2}{c}{Pred. Acc. (\%) ↑} &  & \multicolumn{3}{c}{Spatial DCP (\%) ↑} &  & \multicolumn{3}{c}{Temporal DCP (\%) ↑} \\
\cline{5-6} \cline{8-10} \cline{12-14}
& & &  & Verb & Noun &  & PT & IR & RR &  & PT & IR & RR \\ \hline
Dense & 0 & 7.512 &  & 38.10 & 28.23 &  & 27.20 & 36.04 & 9.41 &  & 96.35 & 41.64 & 33.33 \\ \hline
OMP~\cite{omp}   & \multirow{5}{*}{20} & 6.009 &  & 2.26  & 1.10  &  & \cellcolor[HTML]{FFCCC9}8.05  & 0.39 & 0.07 &  & \cellcolor[HTML]{FFCCC9}6.06  & 0.99 & 0.00 \\
GMP~\cite{gmp}   &                      & 6.879 &  & 7.76  & 0.52  &  & \cellcolor[HTML]{B2F6B1}37.74 & 0.54 & 0.20 &  & \cellcolor[HTML]{FFCCC9}13.21 & 1.07 & 0.07 \\
Kwon et al.~\cite{kwon2022fast} &               & 6.380 &  & 13.75 & 9.41  &  & \cellcolor[HTML]{B2F6B1}30.63 & 11.82 & 3.51 &  & \cellcolor[HTML]{B2F6B1}97.12 & 16.39 & 11.13 \\
ECoFLaP~\cite{ecoflap}   &                  & 5.760 &  & 23.22 & 17.21 &  & \cellcolor[HTML]{B2F6B1}32.01 & 24.95 & 6.96 &  & \cellcolor[HTML]{B2F6B1}98.48 & 25.23 & 21.40 \\
Ours      &                  & 6.593 &  & \textbf{35.93} & \textbf{27.52} &  & \cellcolor[HTML]{B2F6B1}28.64 & \textbf{36.42} & \textbf{9.68} &  & \cellcolor[HTML]{B2F6B1}96.86 & \textbf{39.74} & \textbf{32.74} \\ \hline
OMP~\cite{omp}   & \multirow{5}{*}{25} & 5.633 &  & 2.26  & 1.23  &  & \cellcolor[HTML]{B2F6B1}31.03 & 1.38 & 0.27 &  & \cellcolor[HTML]{FFCCC9}0.00  & 0.00 & 0.00 \\
GMP~\cite{gmp}   &                      & 6.755 &  & 0.42  & 1.00  &  & \cellcolor[HTML]{B2F6B1}33.75 & 0.98 & 0.27 &  & \cellcolor[HTML]{FFCCC9}0.00  & 0.00 & 0.00 \\
Kwon et al.~\cite{kwon2022fast} &               & 6.096 &  & 7.24  & 4.24  &  & \cellcolor[HTML]{FFCCC9}23.02 & 4.06 & 1.19 &  & \cellcolor[HTML]{FFCCC9}93.62 & 8.30 & 4.84 \\
ECoFLaP~\cite{ecoflap}   &                  & 5.765 &  & 23.80 & 17.95 &  & \cellcolor[HTML]{B2F6B1}33.74 & 25.51 & 7.49 &  & \cellcolor[HTML]{B2F6B1}98.51 & 27.80 & 21.87 \\
Ours      &                  & 5.898 &  & \textbf{31.86} & \textbf{22.28} &  & \cellcolor[HTML]{B2F6B1}35.97 & \textbf{33.34} & \textbf{9.68} &  & \cellcolor[HTML]{B2F6B1}97.77 & \textbf{32.09} & \textbf{26.31} \\ \hline
OMP~\cite{omp}   & \multirow{5}{*}{30} & 5.258 &  & 2.26  & 1.26  &  & \cellcolor[HTML]{FFCCC9}21.51 & 1.19 & 0.20 &  & \cellcolor[HTML]{FFCCC9}0.00  & 0.00 & 0.00 \\
GMP~\cite{gmp}   &                      & 6.630 &  & 2.68  & 0.87  &  & \cellcolor[HTML]{FFCCC9}20.59 & 0.26 & 0.07 &  & \cellcolor[HTML]{FFCCC9}60.61 & 2.72 & 0.20 \\
Kwon et al.~\cite{kwon2022fast} &               & 5.813 &  & 2.46  & 0.94  &  & \cellcolor[HTML]{FFCCC9}9.35  & 0.52 & 0.13 &  & \cellcolor[HTML]{FFCCC9}85.61 & 3.86 & 1.19 \\
ECoFLaP~\cite{ecoflap}   &                  & 5.438 &  & 18.60 & 10.25 &  & \cellcolor[HTML]{B2F6B1}32.49 & 15.13 & 4.24 &  & \cellcolor[HTML]{B2F6B1}98.47 & 15.52 & 12.86 \\
Ours      &                  & 5.630 &  & \textbf{28.65} & \textbf{18.95} &  & \cellcolor[HTML]{B2F6B1}33.25 & \textbf{28.75} & \textbf{7.89} &  & \cellcolor[HTML]{B2F6B1}98.31 & \textbf{28.05} & \textbf{23.33} \\ \hline
\end{tabular*}
}
\end{table*}

%% file: tabs/res_actionclip32_ee4d.tex
\begin{table*}[h]
\vspace*{0.1in}
\centering
\caption{
Experimental results with the ActionCLIP ViT-B/32 model on the Ego-Exo4D dataset.
The notation, highlighting scheme, and evaluation protocols are the same as in Table~\ref{tab_res_actionclip32_ekv}.
}
\label{tab_res_actionclip32_ee4d}
{\small                           
\setlength{\tabcolsep}{1pt}      
\renewcommand{\arraystretch}{1.0}

\begin{tabular*}{\textwidth}{@{\extracolsep{\fill}} c c c c c c c c c c c c c}
\hline
\multirow{2}{*}{Pruning} & \multirow{2}{*}{P-Ratio (\%)} & \multirow{2}{*}{GFLOPs} &  & Pred. Acc. (\%) ↑ &  & \multicolumn{3}{c}{Spatial DCP (\%) ↑} &  & \multicolumn{3}{c}{Temporal DCP (\%) ↑} \\
\cline{5-5} \cline{7-9} \cline{11-13}
& & &  & Action &  & PT & IR & RR &  & PT & IR & RR \\ \hline
Dense & 0 & 7.512 &  & 58.83 &  & 16.48 & 60.45 & 10.04 &  & 98.13 & 62.06 & 59.80 \\ \hline
OMP~\cite{omp}   & \multirow{5}{*}{20} & 6.009 &  & 3.25  &  & \cellcolor[HTML]{B2F6B1}86.36 & 5.82 & 1.14 &  & \cellcolor[HTML]{FFCCC9}0.00  & 0.00 & 0.00 \\
GMP~\cite{gmp}   &                      & 6.876 &  & 8.38  &  & \cellcolor[HTML]{FFCCC9}2.48  & 5.43 & 0.36 &  & \cellcolor[HTML]{FFCCC9}18.53 & 11.67 & 2.69 \\
Kwon et al.~\cite{kwon2022fast} &               & 6.380 &  & 36.42 &  & \cellcolor[HTML]{B2F6B1}21.71 & 46.86 & 8.96 &  & \cellcolor[HTML]{FFCCC9}82.78 & 47.96 & 34.17 \\
ECoFLaP~\cite{ecoflap}   &                 & 5.834 &  & 38.76 &  & \cellcolor[HTML]{B2F6B1}19.23 & 47.91 & 9.50 &  & \cellcolor[HTML]{FFCCC9}94.19 & 50.89 & 46.54 \\
Ours      &                 & 6.039 &  & \textbf{49.88} &  & \cellcolor[HTML]{B2F6B1}17.20 & \textbf{51.27} & \textbf{9.68} &  & \cellcolor[HTML]{B2F6B1}98.61 & \textbf{57.96} & \textbf{55.50} \\ \hline
OMP~\cite{omp}   & \multirow{5}{*}{25} & 5.633 &  & 3.40  &  & \cellcolor[HTML]{B2F6B1}78.26 & 5.18 & 1.08 &  & \cellcolor[HTML]{FFCCC9}0.00  & 0.00 & 0.00 \\
GMP~\cite{gmp}   &                      & 6.751 &  & 8.65  &  & \cellcolor[HTML]{FFCCC9}0.43  & 2.87 & 0.06 &  & \cellcolor[HTML]{FFCCC9}40.98 & 24.96 & 5.68 \\
Kwon et al.~\cite{kwon2022fast} &               & 6.096 &  & 23.53 &  & \cellcolor[HTML]{B2F6B1}25.69 & 45.06 & 7.89 &  & \cellcolor[HTML]{FFCCC9}75.48 & 37.09 & 23.18 \\
% ECoFLaP~\cite{ecoflap}   &                 & 5.668 &  & 32.03 &  & \cellcolor[HTML]{B2F6B1}27.73 & 41.42 & \textbf{11.53} &  & \cellcolor[HTML]{FFCCC9}89.08 & 44.96 & 37.04 \\
ECoFLaP~\cite{ecoflap}   &                 & 5.663 &  & 26.29 &  & \cellcolor[HTML]{B2F6B1}23.88 & 39.63 & 7.65 &  & \cellcolor[HTML]{FFCCC9}88.43 & 36.24 & 28.32 \\
Ours      &                 & 5.832 &  & \textbf{45.88} &  & \cellcolor[HTML]{B2F6B1}19.73 & \textbf{50.43} & \textbf{10.63} &  & \cellcolor[HTML]{B2F6B1}98.55 & \textbf{55.71} & \textbf{53.11} \\ \hline
OMP~\cite{omp}   & \multirow{5}{*}{30} & 5.258 &  & 5.20  &  & \cellcolor[HTML]{FFCCC9}0.00  & 0.00 & 0.00 &  & \cellcolor[HTML]{FFCCC9}0.00  & 0.00 & 0.00 \\
GMP~\cite{gmp}   &                      & 6.629 &  & 8.23  &  & \cellcolor[HTML]{FFCCC9}0.00  & 0.00 & 0.00 &  & \cellcolor[HTML]{FFCCC9}3.08  & 6.33 & 0.42 \\
Kwon et al.~\cite{kwon2022fast} &               & 5.813 &  & 13.31 &  & \cellcolor[HTML]{B2F6B1}36.89 & 45.07 & 6.81 &  & \cellcolor[HTML]{FFCCC9}25.57 & 20.74 & 4.72 \\
ECoFLaP~\cite{ecoflap}   &                 & 5.551 &  & 25.63 &  & \cellcolor[HTML]{B2F6B1}28.81 & 35.26 & 9.50 &  & \cellcolor[HTML]{FFCCC9}86.05 & 40.06 & 28.38 \\
Ours      &                 & 5.610 &  & \textbf{44.32} &  & \cellcolor[HTML]{B2F6B1}20.93 & \textbf{48.93} & \textbf{10.93} &  & \cellcolor[HTML]{FFCCC9}97.82 & \textbf{54.36} & \textbf{51.08} \\ \hline
\end{tabular*}
}
\end{table*}

%% file: tex/5_experiment_0204.tex
\input{tabs/abs}

\section{Experiments}
\label{sec_5}

\subsection{Validity of our Method}
We experiment with the same experimental setups in Section~\ref{sec_3_2}.
From Tables~\ref{tab_res_actionclip32_ekv} and~\ref{tab_res_actionclip32_ee4d}, compared with four existing pruning methods, we draw the following observations:
(1) Our method achieves the highest prediction accuracy compared with existing approaches. 
% We interpret this as our method’s ability to better exploit rationale localization for accurate predictions.
(2) Our approach achieves higher IR and RR in most settings, in several cases by a substantial margin.
For example, on the EPIC-KITCHENS VISOR, when the pruning ratio is 30\%, compared with ECoFLaP, our method achieves improvements of 13.62\% and 3.65\% in spatial IR and RR. 
(3) In terms of PT, our method preserves the ability of existing pruning methods to maintain the original model’s PT score, for the most part.
These highlight that \textit{our pruning method not only enables low-latency deployment of VLMs, but also safeguards model reliability by enforcing doubly-correct predictions.} 
Notably, \textit{safeguarding doubly-correct predictions can further enhance prediction accuracy}, as evidenced by our method’s substantial improvement of prediction accuracy over existing works.
We further validate our method in the Appendix with Spatial DCP visualizations, comparisons of layer-wise neuron retention, and additional VLM evaluations.

\subsection{Ablation Studies}
In Table~\ref{tab_abs}, we present ablation studies conducted on the EPIC-KITCHENS VISOR with a pruning ratio of 20\%.

\subsubsection{Effects of Rationale-Informed Pruning}
A key component of our proposed method is the use of rationale information to guide pruning, enabling the model to achieve better doubly-correct predictions. 
When pruning is performed without incorporating rationale information (i.e., without the rationale object mask generated by VLMs), the VLM’s performance of doubly-correct prediction degrades significantly. 
% For instance, Table~\ref{tab_abs} demonstrates that the VLM’s spatial IR decreases by 2.7\%, while its temporal IR decreases by 4.5\%. 
% At the same time, this leads to reductions in the Top-1 Verb and Noun Accuracy by 3.0\% and 3.4\%, respectively.

\subsubsection{Effects of Pruning Mask Weight}
An important design of our method lies in rationale mask construction. 
Instead of directly using the raw rationale mask, we adaptively adjust its semantic factor based on the similarity between the rationale object to be explained and the predicted label. 
This design choice is also critical: when the adaptive semantic factor of the mask is removed, the model’s doubly-correct prediction (DCP) performance exhibits a noticeable decline.
% For example, the model exhibits a 3.1\% decrease in spatial IR and a 4.5\% decrease in temporal IR. 
% Meanwhile, this leads to reductions of 2.9\% and 3.4\% in Top-1 Verb and Noun Accuracy, respectively.

%% file: tabs/abs.tex
\begin{table}[h]
\centering
\caption{
Ablation Studies of our pruning method.
}
% \small
% \resizebox{0.47\textwidth}{!}{%
{\footnotesize                           
\setlength{\tabcolsep}{2.5pt}      
\renewcommand{\arraystretch}{1.0}
\begin{tabular}{ccccccccccccc}
\hline
\multicolumn{2}{c}{Setups}                                                                                         &  & \multicolumn{2}{c}{\begin{tabular}[c]{@{}c@{}}Pred. \\ Acc. (\%) ↑\end{tabular}} &  & \multicolumn{3}{c}{\begin{tabular}[c]{@{}c@{}}Spatial \\ DCP (\%) ↑\end{tabular}} &  & \multicolumn{3}{c}{\begin{tabular}[c]{@{}c@{}}Temporal \\ DCP (\%) ↑\end{tabular}} \\ \cline{1-2} \cline{4-5} \cline{7-9} \cline{11-13} 
\begin{tabular}[c]{@{}c@{}}Rationale \\ mask\end{tabular} & \begin{tabular}[c]{@{}c@{}}Mask \\ Weight\end{tabular} &  & Verb                                   & Noun                                  &  & PT                        & IR                        & RR                      &  & PT                        & IR                        & RR                       \\ \hline
\checkmark                                 & \checkmark                              &  & 35.9                                  & 27.5                                 &  & 28.6                     & 36.4                     & 9.7                    &  & 96.9                     & 39.7                     & 32.7                    \\ \hline
$\times$                                                  & $\times$                                               &  & 32.9                                  & 24.1                                 &  & 33.7                     & 33.7                     & 9.8                    &  & 98.4                     & 35.2                     & 28.6                    \\
\checkmark                                 & $\times$                                               &  & 33.0                                  & 24.1                                 &  & 32.9                     & 33.3                     & 9.5                    &  & 98.4                     & 35.2                     & 28.5                    \\ \hline
\end{tabular}}
\vspace{-0.2cm}
\label{tab_abs}
\end{table}

%% file: tex/6_conclusion.tex
\section{Conclusions}

Preserving doubly-correct predictions is essential for deploying Vision–Language Models (VLMs) that are not only efficient but also safe and trustworthy for egocentric visual understanding.
In this paper, we presented a new study of achieving low-latency VLMs through the lens of doubly-correct prediction.
We showed that existing pruning methods often compromise this property, and proposed a new rationale-informed pruning approach that aligns localized evidence with decision-making to ensure both efficiency and reliability.
On egocentric video benchmarks, we achieve higher inference reliability (IR) and doubly-right samples (RR) than prior methods, without compromising the prediction trustworthiness (PT) of original VLMs.
% Extending the insights of our study to Vision-Language Action (VLA) Models for embodied agents represents a promising direction for future research. 
This work outlines a new paradigm for evaluating efficiency methods, highlighting that true progress must integrate operational safety and human trust with computational efficiency. 
We aim to foster research toward efficient and trustworthy VLMs that bridge accuracy-driven innovation with the low-latency, transparency, and safety needs of responsible embodied intelligence and human-robot collaboration.

%% file: tex/x_appendix.tex
\clearpage

\section*{Appendix}

\begin{figure*}[!b]
  \centering
  \includegraphics[width=1.0\textwidth]{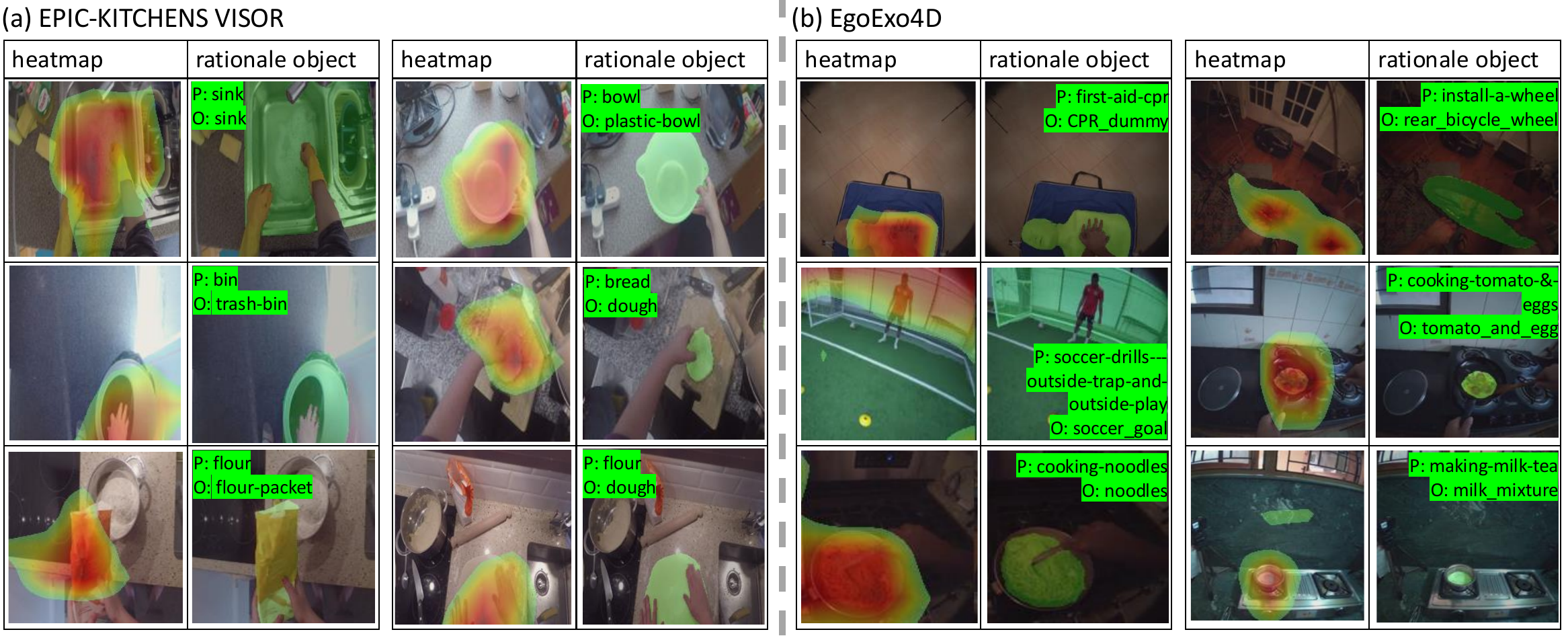}
  \caption{
  % Spatial DCP visualizations of ActionCLIP (ViT-B/32) with our pruning method (with pruning ratio of 20\%).
  Spatial DCP visualization of ActionCLIP (ViT-B/32) under a 20\% pruning ratio with our method.
  % For clearer visualization, we only display the heatmap regions where its pixel values are greater than 0.5.
  For clarity, we display only the heatmap regions with pixel values greater than 0.5.
  The rationale object is overlaid as a green mask.
  We also display the prediction label and the rationale object to be explained in the spatial DCP, highlighted with green background text.
  ``P'' denotes the predicted label and ``O'' denotes the rationale object.
  % Please refer to the Appendix for more visualizations.
  }
  \label{fig_vis}
\end{figure*}

\begin{figure*}[!b]
  \centering
  \includegraphics[width=1.0\textwidth]{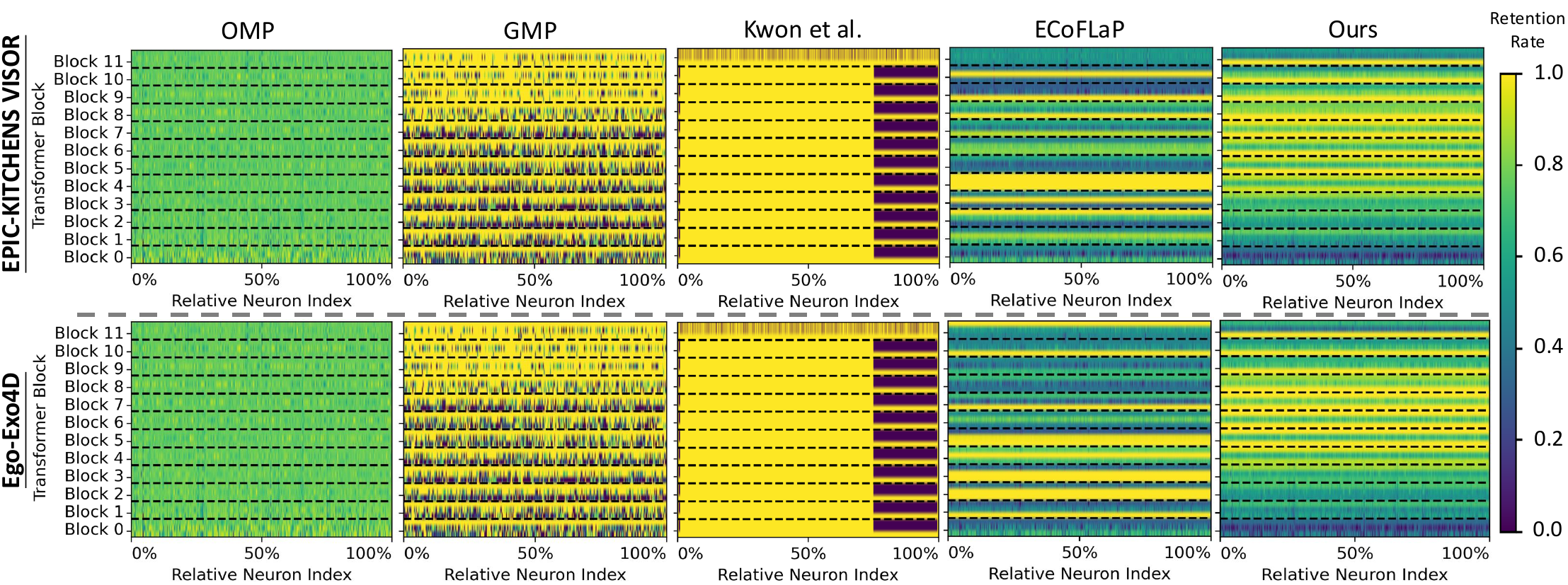}
  \caption{
  Different methods prune different layers and neurons.
  We present the visualization results of ActionCLIP ViT-B/32 with a pruning ratio of 25\%.
  % On the EPIC-KITCHENS VISOR dataset, the results of our method and ECoFLaP have already been presented in Figure~\ref{fig_pruning_vis}.
  }
  \label{fig_pruning_vis_supp}
\end{figure*}

This section provides additional materials to complement the main text. 
It includes:

$\bullet$ Spatial DCP visualization results of the VLM with our pruning methods (see Figure~\ref{fig_vis}).
These demonstrate that the VLM continues to provide valid localization of rationale objects under the spatial DCP evaluation.

$\bullet$ Visualization of layer-wise neuron retention ratios in the Vision Encoder after pruning (see Figure~\ref{fig_pruning_vis_supp}).
The results show that the distinct methods prune different regions of the model. 
Compared with ECoFLaP, which primarily retains neurons in the earlier layers (up to layer 4), our method preserves a higher proportion of neurons in the mid-to-late layers—particularly around layer 6 and 9.
This observation suggests that mid-to-late layers in the Vision Encoder contribute more critically to maintaining model performance.
By emphasizing the retention of deeper semantic representations, our approach better supports doubly-correct prediction, ensuring both accurate outputs and valid rationales, while pruning less influential early layers to achieve lower latency without sacrificing trustworthiness and reliability.

$\bullet$ Quantitative results for ActionCLIP ViT-B/16, CLIP ViT-B/32, and CLIP ViT-B/16 are reported on the EPIC-KITCHENS VISOR (Tables~\ref{tab_res_actionclip16_ekv},~\ref{tab_res_clip32_ekv},~\ref{tab_res_clip16_ekv}) and Ego-Exo4D (Tables~\ref{tab_res_actionclip16_ee4d},~\ref{tab_res_clip32_ee4d},~\ref{tab_res_clip16_ee4d}) datasets.
% These show the same observations as Tables~\ref{tab_res_actionclip32_ekv} and~\ref{tab_res_actionclip32_ee4d} in the main text.
These show observations that are largely consistent with those in Tables~\ref{tab_res_actionclip32_ekv} and~\ref{tab_res_actionclip32_ee4d} in the main text.
Among all six tables, in the spatial and temporal PT columns, scores higher than the dense baseline are highlighted in \highlightg{green}, while the others are highlighted in \highlightr{red}.
In each setup, for prediction accuracy (Pred. Acc.) as well as spatial and temporal IR and RR, we display the first-place results in \textbf{bold}.
The definitions of PT, IR, and RR are provided in Section~\ref{sec_3_1}.
``P-Ratio'' denotes the pruning ratio of the vision encoder of VLMs.
Note that in Table~\ref{tab_res_clip16_ekv}, our method does not surpass the dense baseline in Temporal PT. 
This is primarily because the dense model already achieves 100\% in this metric. 
On the other hand, the drop of our method under this condition is negligible. 
For example, at pruning ratios of 20\%, 25\%, and 30\%, the decreases are only 0.44\%, 0.93\%, and 0.77\%, respectively.

\input{tabs/res_actionclip16_ekv}

\input{tabs/res_clip32_ekv}

\input{tabs/res_clip16_ekv}

Overall, the additional results across ActionCLIP ViT-B/16, CLIP ViT-B/32, and CLIP ViT-B/16 further validate the generality of our method.
Across different backbones and pruning ratios, our method consistently preserves doubly-correct prediction.
Compared with existing weight pruning methods, the gains in IR and RR, together with competitive accuracy and largely preserved PT, show that our method reduces inference cost while maintaining reliability and trustworthiness.
These appendix results further support our conclusion that our conclusion that efficient and safe VLM deployment should consider not only accuracy and latency, but also doubly-correct and trustworthy predictions.

\input{tabs/res_actionclip16_ee4d}

\input{tabs/res_clip32_ee4d}

\input{tabs/res_clip16_ee4d}

%% file: tabs/res_actionclip16_ekv.tex
\begin{table*}[!b]
\centering
{\small
% {
\setlength{\tabcolsep}{4pt}
\renewcommand{\arraystretch}{0.5}

\begin{tabular*}{\textwidth}{@{\extracolsep{\fill}} c c c c c c c c c c c c c c}
\hline
\multirow{2}{*}{Pruning} & \multirow{2}{*}{P-Ratio (\%)} & \multirow{2}{*}{GFLOPs} &  & \multicolumn{2}{c}{Pred. Acc. (\%) ↑} &  & \multicolumn{3}{c}{Spatial DCP (\%) ↑} &  & \multicolumn{3}{c}{Temporal DCP (\%) ↑} \\
\cline{5-6} \cline{8-10} \cline{12-14}
& & &  & Verb & Noun &  & PT & IR & RR &  & PT & IR & RR \\ \hline
Dense   & 0                   & 24.159 &  & 42.82 & 32.21 &  & 33.91  & 38.13 & 12.99 &  & 98.44 & 45.59 & 37.71 \\ \hline
OMP~\cite{omp}     & \multirow{5}{*}{20} & 19.326 &  & 3.36  & 0.52  &  & \cellcolor[HTML]{FFCCC9}0.00   & 0.00  & 0.00  &  & \cellcolor[HTML]{FFCCC9}0.00  & 0.00  & 0.00  \\
GMP~\cite{gmp}     &                     & 21.708 &  & 2.30  & 1.03  &  & \cellcolor[HTML]{B2F6B1}44.44  & 0.88  & 0.27  &  & \cellcolor[HTML]{FFCCC9}55.56 & 2.07  & 0.33  \\
Kwon et al.~\cite{kwon2022fast}    &                     & 19.700 &  & 23.64 & 14.10 &  & \cellcolor[HTML]{FFCCC9}29.86  & 18.53 & 5.50  &  & \cellcolor[HTML]{FFCCC9}97.12 & 24.66 & 17.89 \\
ECoFLaP~\cite{ecoflap} &                     & 17.713 &  & 24.32 & 14.78 &  & \cellcolor[HTML]{B2F6B1}39.86  & 17.78 & 7.42  &  & \cellcolor[HTML]{FFCCC9}98.22 & 27.68 & 18.29 \\
Ours    &                     & 18.511 &  & \textbf{36.87} & \textbf{26.68} &  & \cellcolor[HTML]{B2F6B1}37.58  & \textbf{31.75} & \textbf{11.93} &  & \cellcolor[HTML]{B2F6B1}99.37 & \textbf{36.56} & \textbf{31.54} \\ \hline
OMP~\cite{omp}     & \multirow{5}{*}{25} & 18.117 &  & 0.23  & 0.52  &  & \cellcolor[HTML]{B2F6B1}100.00 & 0.27  & 0.07  &  & \cellcolor[HTML]{FFCCC9}0.00  & 0.00  & 0.00  \\
GMP~\cite{gmp}     &                     & 21.225 &  & 0.06  & 0.65  &  & \cellcolor[HTML]{FFCCC9}0.00   & 0.00  & 0.00  &  & \cellcolor[HTML]{FFCCC9}0.00  & 0.00  & 0.00  \\
Kwon et al.~\cite{kwon2022fast}    &                     & 18.582 &  & 8.25  & 4.20  &  & \cellcolor[HTML]{B2F6B1}41.33  & 6.65  & 2.05  &  & \cellcolor[HTML]{FFCCC9}97.33 & 8.83  & 4.84  \\
ECoFLaP~\cite{ecoflap} &                     & 15.476 &  & 8.51  & 4.72  &  & \cellcolor[HTML]{B2F6B1}36.90  & 6.10  & 2.05  &  & \cellcolor[HTML]{FFCCC9}92.86 & 12.40 & 5.17  \\
Ours    &                     & 17.650 &  & \textbf{33.76} & \textbf{24.09} &  & \cellcolor[HTML]{B2F6B1}37.70  & \textbf{28.11} & \textbf{11.07} &  & \cellcolor[HTML]{B2F6B1}99.32 & \textbf{34.06} & \textbf{29.16} \\ \hline
OMP~\cite{omp}     & \multirow{5}{*}{30} & 16.909 &  & 0.10  & 0.36  &  & \cellcolor[HTML]{FFCCC9}0.00   & 0.00  & 0.00  &  & \cellcolor[HTML]{FFCCC9}0.00  & 0.00  & 0.00  \\
GMP~\cite{gmp}     &                     & 20.719 &  & 1.58  & 1.07  &  & \cellcolor[HTML]{FFCCC9}25.00  & 0.82  & 0.20  &  & \cellcolor[HTML]{FFCCC9}0.00  & 0.00  & 0.00  \\
Kwon et al.~\cite{kwon2022fast}    &                     & 17.463 &  & 0.49  & 1.33  &  & \cellcolor[HTML]{FFCCC9}30.77  & 2.16  & 0.53  &  & \cellcolor[HTML]{FFCCC9}96.15 & 4.73  & 1.66  \\
ECoFLaP~\cite{ecoflap} &                     & 15.001 &  & 11.71 & 4.46  &  & \cellcolor[HTML]{FFCCC9}32.88  & 5.02  & 1.59  &  & \cellcolor[HTML]{FFCCC9}94.52 & 9.69  & 4.57  \\
Ours    &                     & 16.661 &  & \textbf{31.89} & \textbf{21.47} &  & \cellcolor[HTML]{B2F6B1}35.79  & \textbf{24.19} & \textbf{9.34}  &  & \cellcolor[HTML]{B2F6B1}99.49 & \textbf{30.25} & \textbf{25.98} \\ \hline
\end{tabular*}
}
\caption{Experimental results with the ActionCLIP ViT-B/16 model on the EPIC-KITCHENS VISOR dataset.
}
\label{tab_res_actionclip16_ekv}
\end{table*}

%% file: tabs/res_clip32_ekv.tex
\begin{table*}[!b]
\centering
{\small
% {
\setlength{\tabcolsep}{4pt}
\renewcommand{\arraystretch}{0.5}

\begin{tabular*}{\textwidth}{@{\extracolsep{\fill}} c c c c c c c c c c c c c c}
\hline
\multirow{2}{*}{Pruning} & \multirow{2}{*}{P-Ratio (\%)} & \multirow{2}{*}{GFLOPs} &  & \multicolumn{2}{c}{Pred. Acc. (\%) ↑} &  & \multicolumn{3}{c}{Spatial DCP (\%) ↑} &  & \multicolumn{3}{c}{Temporal DCP (\%) ↑} \\
\cline{5-6} \cline{8-10} \cline{12-14}
& & &  & Verb & Noun &  & PT & IR & RR &  & PT & IR & RR \\ \hline
Dense   & 0                   & 7.512 &  & 30.76 & 26.65 &  & 10.80 & 40.30 & 3.58 &  & 98.80  & 37.40 & 32.74 \\ \hline
OMP~\cite{omp}     & \multirow{5}{*}{20} & 6.009 &  & 3.07  & 0.00  &  & \cellcolor[HTML]{FFCCC9}0.00  & 0.00  & 0.00 &  & \cellcolor[HTML]{FFCCC9}0.00   & 0.00  & 0.00  \\
GMP~\cite{gmp}     &                     & 6.878 &  & 14.59 & 1.78  &  & \cellcolor[HTML]{FFCCC9}0.00  & 0.00  & 0.00 &  & \cellcolor[HTML]{B2F6B1}100.00 & 4.75  & 1.06  \\
Kwon et al.~\cite{kwon2022fast}    &     & 6.380 &  & 13.58 & 9.38  &  & \cellcolor[HTML]{FFCCC9}9.55  & 9.71  & 1.13 &  & \cellcolor[HTML]{B2F6B1}99.44  & 16.15 & 11.73 \\
ECoFLaP~\cite{ecoflap} &                 & 5.901 &  & 16.04 & 14.00 &  & \cellcolor[HTML]{FFCCC9}8.70  & 16.30 & 1.46 &  & \cellcolor[HTML]{B2F6B1}99.21  & 18.94 & 16.63 \\
Ours    &                                & 6.117 &  & \textbf{26.78} & \textbf{21.96} &  & \cellcolor[HTML]{FFCCC9}10.23 & \textbf{33.33} & \textbf{2.92} &  & \cellcolor[HTML]{B2F6B1}99.07  & \textbf{33.15} & \textbf{28.23} \\ \hline
OMP~\cite{omp}     & \multirow{5}{*}{25} & 5.633 &  & 3.53  & 0.29  &  & \cellcolor[HTML]{FFCCC9}0.00  & 0.00  & 0.00 &  & \cellcolor[HTML]{B2F6B1}100.00 & 0.56  & 0.13  \\
GMP~\cite{gmp}     &                     & 6.754 &  & 6.18  & 1.16  &  & \cellcolor[HTML]{FFCCC9}4.76  & 0.39  & 0.07 &  & \cellcolor[HTML]{B2F6B1}100.00 & 4.75  & 1.06  \\
Kwon et al.~\cite{kwon2022fast}    &     & 6.096 &  & 11.19 & 4.46  &  & \cellcolor[HTML]{FFCCC9}4.11  & 1.55  & 0.20 &  & \cellcolor[HTML]{FFCCC9}97.26  & 7.68  & 4.71  \\
ECoFLaP~\cite{ecoflap} &                 & 5.776 &  & 14.46 & 13.32 &  & \cellcolor[HTML]{B2F6B1}12.20 & 23.85 & 2.05 &  & \cellcolor[HTML]{B2F6B1}99.61  & 19.28 & 16.77 \\
Ours    &                                & 5.879 &  & \textbf{25.03} & \textbf{19.40} &  & \cellcolor[HTML]{B2F6B1}11.02 & \textbf{28.97} & \textbf{2.78} &  & \cellcolor[HTML]{B2F6B1}99.74  & \textbf{29.34} & \textbf{25.18} \\ \hline
OMP~\cite{omp}     & \multirow{5}{*}{30} & 5.258 &  & 4.14  & 0.91  &  & \cellcolor[HTML]{FFCCC9}10.00 & 0.35  & 0.07 &  & \cellcolor[HTML]{FFCCC9}70.00  & 2.55  & 0.46  \\
GMP~\cite{gmp}     &                     & 6.631 &  & 16.24 & 1.20  &  & \cellcolor[HTML]{FFCCC9}4.35  & 0.78  & 0.07 &  & \cellcolor[HTML]{FFCCC9}65.22  & 9.26  & 0.99  \\
Kwon et al.~\cite{kwon2022fast}    &     & 5.813 &  & 13.29 & 2.65  &  & \cellcolor[HTML]{FFCCC9}6.06  & 1.23  & 0.13 &  & \cellcolor[HTML]{FFCCC9}93.94  & 4.61  & 2.05  \\
ECoFLaP~\cite{ecoflap} &                 & 5.504 &  & 10.64 & 9.83  &  & \cellcolor[HTML]{FFCCC9}10.50 & 9.69  & 1.26 &  & \cellcolor[HTML]{B2F6B1}98.90  & 14.88 & 11.86 \\
Ours    &                                & 5.620 &  & \textbf{20.70} & \textbf{15.56} &  & \cellcolor[HTML]{B2F6B1}13.87 & \textbf{24.43} & \textbf{2.85} &  & \cellcolor[HTML]{B2F6B1}100.00 & \textbf{23.92} & \textbf{20.54} \\ \hline
\end{tabular*}
}
\caption{Experimental results with the CLIP ViT-B/32 model on the EPIC-KITCHENS VISOR dataset.
}
\label{tab_res_clip32_ekv}
\end{table*}

%% file: tabs/res_clip16_ekv.tex
\begin{table*}[!b]
\centering
{\small
% {
\setlength{\tabcolsep}{4pt}
\renewcommand{\arraystretch}{0.5}

\begin{tabular*}{\textwidth}{@{\extracolsep{\fill}} c c c c c c c c c c c c c c}
\hline
\multirow{2}{*}{Pruning} & \multirow{2}{*}{P-Ratio (\%)} & \multirow{2}{*}{GFLOPs} &  & \multicolumn{2}{c}{Pred. Acc. (\%) ↑} &  & \multicolumn{3}{c}{Spatial DCP (\%) ↑} &  & \multicolumn{3}{c}{Temporal DCP (\%) ↑} \\
\cline{5-6} \cline{8-10} \cline{12-14}
& & &  & Verb & Noun &  & PT & IR & RR &  & PT & IR & RR \\ \hline
Dense   & 0                   & 24.159 &  & 34.38 & 31.82 &  & 2.93  & 32.69 & 1.13 &  & 100.00 & 42.01 & 38.50 \\ \hline
OMP~\cite{omp}     & \multirow{5}{*}{20} & 19.326 &  & 16.88 & 0.81  &  & \cellcolor[HTML]{FFCCC9}0.00  & 0.00  & 0.00 &  & \cellcolor[HTML]{FFCCC9}100.00 & 0.51  & 0.27  \\
GMP~\cite{gmp}     &                     & 21.702 &  & 17.69 & 2.72  &  & \cellcolor[HTML]{B2F6B1}3.57  & 0.79  & 0.07 &  & \cellcolor[HTML]{FFCCC9}100.00 & 9.49  & 1.86  \\
Kwon et al.~\cite{kwon2022fast}    &                     & 19.700 &  & 13.13 & 13.97 &  & \cellcolor[HTML]{FFCCC9}2.13  & 17.65 & 0.40 &  & \cellcolor[HTML]{FFCCC9}97.87  & 21.87 & 18.29 \\
ECoFLaP~\cite{ecoflap} &                     & 16.808 &  & 23.93 & \textbf{25.32} &  & \cellcolor[HTML]{B2F6B1}6.84  & \textbf{38.64} & 2.25 &  & \cellcolor[HTML]{FFCCC9}99.80  & \textbf{36.88} & \textbf{32.87} \\
Ours    &                     & 18.420 &  & \textbf{29.24} & 24.26 &  & \cellcolor[HTML]{B2F6B1}8.17  & 35.24 & \textbf{2.45} &  & \cellcolor[HTML]{FFCCC9}99.56  & 34.56 & 29.89 \\ \hline
OMP~\cite{omp}     & \multirow{5}{*}{25} & 18.117 &  & 25.19 & 0.84  &  & \cellcolor[HTML]{FFCCC9}0.00  & 0.00  & 0.00 &  & \cellcolor[HTML]{FFCCC9}75.00  & 0.58  & 0.20  \\
GMP~\cite{gmp}     &                     & 21.215 &  & 19.40 & 2.10  &  & \cellcolor[HTML]{B2F6B1}3.70  & 2.27  & 0.07 &  & \cellcolor[HTML]{FFCCC9}88.89  & 12.83 & 1.59  \\
Kwon et al.~\cite{kwon2022fast}    &                     & 18.582 &  & 2.20  & 4.33  &  & \cellcolor[HTML]{B2F6B1}3.45  & 15.00 & 0.20 &  & \cellcolor[HTML]{FFCCC9}98.85  & 9.01  & 5.70  \\
ECoFLaP~\cite{ecoflap} &                     & 15.872 &  & 20.70 & 22.25 &  & \cellcolor[HTML]{B2F6B1}5.97  & 33.78 & 1.66 &  & \cellcolor[HTML]{FFCCC9}99.76  & 32.13 & 27.70 \\
Ours    &                     & 17.553 &  & \textbf{26.16} & \textbf{22.67} &  & \cellcolor[HTML]{B2F6B1}9.26  & \textbf{35.09} & \textbf{2.65} &  & \cellcolor[HTML]{FFCCC9}99.07  & \textbf{33.52} & \textbf{28.36} \\ \hline
OMP~\cite{omp}     & \multirow{5}{*}{30} & 16.909 &  & 8.67  & 0.94  &  & \cellcolor[HTML]{FFCCC9}0.00  & 0.00  & 0.00 &  & \cellcolor[HTML]{FFCCC9}85.71  & 1.22  & 0.40  \\
GMP~\cite{gmp}     &                     & 20.708 &  & 17.79 & 1.97  &  & \cellcolor[HTML]{FFCCC9}0.00  & 0.00  & 0.00 &  & \cellcolor[HTML]{FFCCC9}96.15  & 18.38 & 1.66  \\
Kwon et al.~\cite{kwon2022fast}    &                     & 17.463 &  & 0.36  & 0.87  &  & \cellcolor[HTML]{FFCCC9}0.00  & 0.00  & 0.00 &  & \cellcolor[HTML]{FFCCC9}100.00 & 2.77  & 1.13  \\
ECoFLaP~\cite{ecoflap} &                     & 15.605 &  & 13.23 & 16.27 &  & \cellcolor[HTML]{B2F6B1}4.84  & 25.86 & 0.99 &  & \cellcolor[HTML]{FFCCC9}99.03  & 25.91 & 20.34 \\
Ours    &                     & 16.580 &  & \textbf{23.16} & \textbf{20.67} &  & \cellcolor[HTML]{B2F6B1}11.00 & \textbf{27.92} & \textbf{2.85} &  & \cellcolor[HTML]{FFCCC9}99.23  & \textbf{31.21} & \textbf{25.71} \\ \hline
\end{tabular*}
}
\caption{Experimental results with the CLIP ViT-B/16 model on the EPIC-KITCHENS VISOR dataset.
}
\label{tab_res_clip16_ekv}
\end{table*}

%% file: tabs/res_actionclip16_ee4d.tex
\begin{table*}[h]
\centering
{\small                           
\setlength{\tabcolsep}{4pt}      
\renewcommand{\arraystretch}{0.8}

\begin{tabular*}{\textwidth}{@{\extracolsep{\fill}} c c c c c c c c c c c c c}
\hline
\multirow{2}{*}{Pruning} & \multirow{2}{*}{P-Ratio (\%)} & \multirow{2}{*}{GFLOPs} &  & Pred. Acc. (\%) ↑ &  & \multicolumn{3}{c}{Spatial DCP (\%) ↑} &  & \multicolumn{3}{c}{Temporal DCP (\%) ↑} \\
\cline{5-5} \cline{7-9} \cline{11-13}
& & &  & Action &  & PT & IR & RR &  & PT & IR & RR \\ \hline
Dense   & 0                   & 24.159 &  & 63.97 &  & 7.28  & 75.70 & 4.84 &  & 96.86 & 68.58  & 64.40 \\ \hline
OMP~\cite{omp}     & \multirow{5}{*}{20} & 19.326 &  & 5.95  &  & \cellcolor[HTML]{B2F6B1}16.98 & 5.16  & 1.08 &  & \cellcolor[HTML]{FFCCC9}6.60  & 41.18  & 0.42  \\
GMP~\cite{gmp}     &                     & 21.704 &  & 6.28  &  & \cellcolor[HTML]{B2F6B1}3.33  & 0.68  & 0.06 &  & \cellcolor[HTML]{FFCCC9}3.33  & 100.00 & 0.06  \\
Kwon et al.~\cite{kwon2022fast}    &                     & 19.700 &  & 44.68 &  & \cellcolor[HTML]{B2F6B1}14.17 & 57.08 & \textbf{7.23} &  & \cellcolor[HTML]{FFCCC9}88.52 & 57.84  & 45.16 \\
ECoFLaP~\cite{ecoflap} &                     & 17.344 &  & 54.48 &  & \cellcolor[HTML]{B2F6B1}9.76  & 65.33 & 5.85 &  & \cellcolor[HTML]{B2F6B1}98.51 & 61.01  & 59.08 \\
Ours    &                     & 18.278 &  & \textbf{58.23} &  & \cellcolor[HTML]{B2F6B1}8.50  & \textbf{65.91} & 5.20 &  & \cellcolor[HTML]{B2F6B1}98.53 & \textbf{63.80}  & \textbf{60.22} \\ \hline
OMP~\cite{omp}     & \multirow{5}{*}{25} & 18.117 &  & 5.77  &  & \cellcolor[HTML]{B2F6B1}15.25 & 4.27  & 0.54 &  & \cellcolor[HTML]{FFCCC9}0.00  & 0.00   & 0.00  \\
GMP~\cite{gmp}     &                     & 21.215 &  & 6.13  &  & \cellcolor[HTML]{FFCCC9}0.00  & 0.00  & 0.00 &  & \cellcolor[HTML]{FFCCC9}0.00  & 0.00   & 0.00  \\
Kwon et al.~\cite{kwon2022fast}    &                     & 18.582 &  & 22.90 &  & \cellcolor[HTML]{B2F6B1}20.56 & 37.77 & 5.26 &  & \cellcolor[HTML]{FFCCC9}81.07 & 31.98  & 20.73 \\
ECoFLaP~\cite{ecoflap} &                     & 16.899 &  & 50.69 &  & \cellcolor[HTML]{B2F6B1}10.29 & \textbf{68.97} & 5.97 &  & \cellcolor[HTML]{B2F6B1}98.77 & 59.30  & 57.35 \\
Ours    &                     & 17.461 &  & \textbf{55.47} &  & \cellcolor[HTML]{B2F6B1}10.51 & 67.74 & \textbf{6.27} &  & \cellcolor[HTML]{B2F6B1}98.00 & \textbf{62.32}  & \textbf{58.48} \\ \hline
OMP~\cite{omp}     & \multirow{5}{*}{30} & 16.910 &  & 5.89  &  & \cellcolor[HTML]{B2F6B1}23.08 & 4.99  & 1.08 &  & \cellcolor[HTML]{FFCCC9}6.41  & 10.20  & 0.30  \\
GMP~\cite{gmp}     &                     & 6.730  &  & 22.87 &  & \cellcolor[HTML]{FFCCC9}0.00  & 0.00  & 1.25 &  & \cellcolor[HTML]{FFCCC9}0.00  & 0.00   & 0.00  \\
Kwon et al.~\cite{kwon2022fast}    &                     & 17.463 &  & 10.22 &  & \cellcolor[HTML]{B2F6B1}33.91 & 23.17 & 4.72 &  & \cellcolor[HTML]{FFCCC9}78.97 & 21.90  & 10.99 \\
ECoFLaP~\cite{ecoflap} &                     & 16.264 &  & 48.59 &  & \cellcolor[HTML]{B2F6B1}11.57 & 59.24 & 6.51 &  & \cellcolor[HTML]{B2F6B1}98.51 & 57.25  & \textbf{55.44} \\
Ours    &                     & 16.596 &  & \textbf{51.44} &  & \cellcolor[HTML]{B2F6B1}12.53 & \textbf{64.64} & \textbf{6.99} &  & \cellcolor[HTML]{B2F6B1}98.61 & \textbf{59.38}  & 55.02 \\ \hline
\end{tabular*}
}
\caption{
Experimental results with the ActionCLIP ViT-B/16 model on the Ego-Exo4D dataset.
}
\label{tab_res_actionclip16_ee4d}
\end{table*}

%% file: tabs/res_clip32_ee4d.tex
\begin{table*}[h]
\centering
{\small                           
\setlength{\tabcolsep}{4pt}      
\renewcommand{\arraystretch}{0.8}

\begin{tabular*}{\textwidth}{@{\extracolsep{\fill}} c c c c c c c c c c c c c}
\hline
\multirow{2}{*}{Pruning} & \multirow{2}{*}{P-Ratio (\%)} & \multirow{2}{*}{GFLOPs} &  & Pred. Acc. (\%) ↑ &  & \multicolumn{3}{c}{Spatial DCP (\%) ↑} &  & \multicolumn{3}{c}{Temporal DCP (\%) ↑} \\
\cline{5-5} \cline{7-9} \cline{11-13}
& & &  & Action &  & PT & IR & RR &  & PT & IR & RR \\ \hline
Dense   & 0                   & 7.512 &  & 62.44 &  & 14.11 & 70.89 & 9.02  &  & 98.69  & 64.83 & 63.08 \\ \hline
OMP~\cite{omp}     & \multirow{5}{*}{20} & 6.009 &  & 1.05  &  & \cellcolor[HTML]{FFCCC9}12.00 & 0.95  & 0.18  &  & \cellcolor[HTML]{FFCCC9}96.00  & 5.25  & 1.43  \\
GMP~\cite{gmp}     &                     & 6.875 &  & 2.88  &  & \cellcolor[HTML]{B2F6B1}21.62 & 4.88  & 0.96  &  & \cellcolor[HTML]{FFCCC9}79.73  & 13.11 & 3.52  \\
Kwon et al.~\cite{kwon2022fast}    &                     & 6.380 &  & 30.02 &  & \cellcolor[HTML]{B2F6B1}22.01 & 43.12 & 6.93  &  & \cellcolor[HTML]{FFCCC9}93.17  & 36.29 & 29.33 \\
ECoFLaP~\cite{ecoflap} &                     & 5.903 &  & 42.22 &  & \cellcolor[HTML]{B2F6B1}16.86 & \textbf{67.68} & 8.00  &  & \cellcolor[HTML]{B2F6B1}100.00 & 50.32 & 47.49 \\
Ours    &                     & 5.962 &  & \textbf{50.87} &  & \cellcolor[HTML]{B2F6B1}21.96 & 66.56 & \textbf{11.89} &  & \cellcolor[HTML]{B2F6B1}100.00 & \textbf{55.45} & \textbf{54.12} \\ \hline
OMP~\cite{omp}     & \multirow{5}{*}{25} & 5.633 &  & 2.13  &  & \cellcolor[HTML]{B2F6B1}14.58 & 2.11  & 0.42  &  & \cellcolor[HTML]{B2F6B1}100.00 & 7.19  & 2.87  \\
GMP~\cite{gmp}     &                     & 6.750 &  & 3.82  &  & \cellcolor[HTML]{FFCCC9}8.47  & 2.35  & 0.60  &  & \cellcolor[HTML]{FFCCC9}72.88  & 12.97 & 5.14  \\
Kwon et al.~\cite{kwon2022fast}    &                     & 6.096 &  & 22.87 &  & \cellcolor[HTML]{B2F6B1}15.72 & 26.29 & 3.64  &  & \cellcolor[HTML]{FFCCC9}91.49  & 31.42 & 21.21 \\
ECoFLaP~\cite{ecoflap} &                     & 5.693 &  & 35.43 &  & \cellcolor[HTML]{FFCCC9}13.52 & 53.93 & 5.73  &  & \cellcolor[HTML]{FFCCC9}98.59  & 47.75 & 41.82 \\
Ours    &                     & 5.760 &  & \textbf{45.85} &  & \cellcolor[HTML]{B2F6B1}21.72 & \textbf{65.85} & \textbf{11.29} &  & \cellcolor[HTML]{B2F6B1}100.00 & \textbf{53.70} & \textbf{51.97} \\ \hline
OMP~\cite{omp}     & \multirow{5}{*}{30} & 5.257 &  & 0.48  &  & \cellcolor[HTML]{FFCCC9}0.00  & 0.00  & 0.00  &  & \cellcolor[HTML]{FFCCC9}0.00   & 0.00  & 0.00  \\
GMP~\cite{gmp}     &                     & 6.629 &  & 1.65  &  & \cellcolor[HTML]{FFCCC9}0.00  & 0.00  & 0.00  &  & \cellcolor[HTML]{B2F6B1}100.00 & 5.07  & 2.51  \\
Kwon et al.~\cite{kwon2022fast}    &                     & 5.813 &  & 9.98  &  & \cellcolor[HTML]{FFCCC9}5.00  & 6.90  & 0.36  &  & \cellcolor[HTML]{FFCCC9}65.00  & 20.53 & 4.66  \\
ECoFLaP~\cite{ecoflap} &                     & 5.428 &  & 37.05 &  & \cellcolor[HTML]{FFCCC9}12.57 & 62.25 & 5.62  &  & \cellcolor[HTML]{B2F6B1}99.20  & \textbf{49.77} & 44.32 \\
Ours    &                     & 5.551 &  & \textbf{39.96} &  & \cellcolor[HTML]{B2F6B1}23.36 & \textbf{65.14} & \textbf{11.05} &  & \cellcolor[HTML]{B2F6B1}99.37  & 49.37 & \textbf{47.01} \\ \hline
\end{tabular*}
}
\caption{
Experimental results with the CLIP ViT-B/32 model on the Ego-Exo4D dataset.
}
\label{tab_res_clip32_ee4d}
\end{table*}

%% file: tabs/res_clip16_ee4d.tex
\begin{table*}[h]
\centering
{\small                           
\setlength{\tabcolsep}{4pt}      
\renewcommand{\arraystretch}{0.8}

\begin{tabular*}{\textwidth}{@{\extracolsep{\fill}} c c c c c c c c c c c c c}
\hline
\multirow{2}{*}{Pruning} & \multirow{2}{*}{P-Ratio (\%)} & \multirow{2}{*}{GFLOPs} &  & Pred. Acc. (\%) ↑ &  & \multicolumn{3}{c}{Spatial DCP (\%) ↑} &  & \multicolumn{3}{c}{Temporal DCP (\%) ↑} \\
\cline{5-5} \cline{7-9} \cline{11-13}
& & &  & Action &  & PT & IR & RR &  & PT & IR & RR \\ \hline
Dense   & 0                   & 24.159 &  & 64.06 &  & 0.55 & 85.71  & 0.36 &  & 97.35  & 66.36 & 63.62 \\ \hline
OMP~\cite{omp}     & \multirow{5}{*}{20} & 19.326 &  & 0.48  &  & \cellcolor[HTML]{FFCCC9}0.00 & 0.00   & 0.00 &  & \cellcolor[HTML]{FFCCC9}0.00   & 0.00  & 0.00  \\
GMP~\cite{gmp}     &                     & 21.697 &  & 1.98  &  & \cellcolor[HTML]{FFCCC9}0.00 & 0.00   & 0.00 &  & \cellcolor[HTML]{B2F6B1}100.00 & 75.00 & 0.36  \\
Kwon et al.~\cite{kwon2022fast}    &                     & 19.700 &  & 32.09 &  & \cellcolor[HTML]{FFCCC9}0.00 & 0.00   & 0.00 &  & \cellcolor[HTML]{FFCCC9}96.97  & 43.69 & 38.23 \\
% ECoFLaP~\cite{ecoflap} &                     & 17.945 &  & 49.88 &  & \cellcolor[HTML]{B2F6B1}3.46 & 37.04  & \textbf{1.79} &  & \cellcolor[HTML]{B2F6B1}99.88  & 53.56 & 51.73 \\
ECoFLaP~\cite{ecoflap} &                     & 17.704 &  & 37.86 &  & \cellcolor[HTML]{FFCCC9}0.27 & 25.00  & 0.12 &  & \cellcolor[HTML]{B2F6B1}99.87  & 46.76 & 44.80 \\
Ours    &                     & 18.050 &  & \textbf{55.05} &  & \cellcolor[HTML]{B2F6B1}1.76 & \textbf{51.52}  & \textbf{1.02} &  & \cellcolor[HTML]{B2F6B1}99.38  & \textbf{59.38} & \textbf{57.47} \\ \hline
OMP~\cite{omp}     & \multirow{5}{*}{25} & 18.117 &  & 0.48  &  & \cellcolor[HTML]{FFCCC9}0.00 & 0.00   & 0.00 &  & \cellcolor[HTML]{FFCCC9}0.00   & 0.00  & 0.00  \\
GMP~\cite{gmp}     &                     & 21.203 &  & 0.48  &  & \cellcolor[HTML]{FFCCC9}0.00 & 0.00   & 0.00 &  & \cellcolor[HTML]{FFCCC9}0.00   & 0.00  & 0.00  \\
Kwon et al.~\cite{kwon2022fast}    &                     & 18.582 &  & 9.98  &  & \cellcolor[HTML]{FFCCC9}0.40 & \textbf{100.00} & 0.06 &  & \cellcolor[HTML]{FFCCC9}92.34  & 25.88 & 13.68 \\
ECoFLaP~\cite{ecoflap} &                     & 16.784 &  & 35.61 &  & \cellcolor[HTML]{B2F6B1}1.72 & 33.33  & 0.78 &  & \cellcolor[HTML]{B2F6B1}98.41  & 51.60 & 44.32 \\
Ours    &                     & 17.230 &  & \textbf{47.27} &  & \cellcolor[HTML]{B2F6B1}2.84 & 55.56  & \textbf{1.49} &  & \cellcolor[HTML]{B2F6B1}99.32  & \textbf{54.52} & \textbf{52.27} \\ \hline
OMP~\cite{omp}     & \multirow{5}{*}{30} & 16.908 &  & 0.48  &  & \cellcolor[HTML]{FFCCC9}0.00 & 0.00   & 0.00 &  & \cellcolor[HTML]{FFCCC9}0.00   & 0.00  & 0.00  \\
GMP~\cite{gmp}     &                     & 20.706 &  & 0.48  &  & \cellcolor[HTML]{FFCCC9}0.00 & 0.00   & 0.00 &  & \cellcolor[HTML]{FFCCC9}0.00   & 0.00  & 0.00  \\
Kwon et al.~\cite{kwon2022fast}    &                     & 17.463 &  & 0.66  &  & \cellcolor[HTML]{FFCCC9}0.00 & 0.00   & 0.00 &  & \cellcolor[HTML]{B2F6B1}100.00 & 29.58 & 1.25  \\
ECoFLaP~\cite{ecoflap} &                     & 16.349 &  & 27.07 &  & \cellcolor[HTML]{B2F6B1}2.45 & \textbf{60.00}  & 0.90 &  & \cellcolor[HTML]{FFCCC9}95.74  & 49.20 & 34.95 \\
Ours    &                     & 16.414 &  & \textbf{40.26} &  & \cellcolor[HTML]{B2F6B1}3.12 & 46.30  & \textbf{1.49} &  & \cellcolor[HTML]{B2F6B1}98.88  & \textbf{50.22} & \textbf{47.25} \\ \hline
\end{tabular*}
}
\caption{
Experimental results with the CLIP ViT-B/16 model on the Ego-Exo4D dataset.
}
\label{tab_res_clip16_ee4d}
\end{table*}